\documentclass{article}
\usepackage[utf8]{inputenc}
\usepackage{authblk}
\usepackage{setspace}
\usepackage[margin=1.25in]{geometry}
\usepackage{graphicx}
\graphicspath{ {./figures/} }
\usepackage{subcaption}
\usepackage{amsmath}
\usepackage[backend=bibtex,style=nejm,citestyle=numeric-comp,sorting=none]{biblatex}
\title{CrystalTac: Vision-Based Tactile Sensor Family Fabricated Via Rapid Monolithic Manufacturing}

\author[1$\dag$]{Wen Fan}
\author[2$\dag$]{Haoran Li}
\author[1*]{Dandan Zhang}

\affil[1]{Department of Bioengineering, Imperial College London, London, UK.}
\affil[2]{School of Engineering Mathematics and Technology, Bristol Robotics Lab, University of Bristol, Bristol, UK.}
\affil[*]{Address correspondence to: d.zhang17@imperial.ac.uk}
\affil[$\dag$]{These authors contributed equally to this work.}

\date{}

\begin{document}

\maketitle

%%%%%% Abstract %%%%%%
\begin{abstract}

Recently, vision-based tactile sensors (VBTSs) have gained popularity in robotics systems. 
The sensing mechanisms of most VBTSs can be categorised based on the type of tactile features they capture.  Each category requires specific structural designs to convert physical contact into optical information. The complex architectures of VBTSs pose challenges for traditional manufacturing techniques in terms of design flexibility, cost-effectiveness, and quality stability. Previous research has shown that monolithic manufacturing using multi-material 3D printing technology can partially address these challenges. This study introduces the CrystalTac family, a series of VBTSs designed with a unique sensing mechanism and fabricated through rapid monolithic manufacturing. Case studies on CrystalTac-type sensors demonstrate their effective performance in tasks involving tactile perception, along with impressive cost-effectiveness and design flexibility. The CrystalTac family aims to highlight the potential of monolithic manufacturing in VBTS development and inspire further research in tactile sensing and manipulation.

% Vision-based tactile sensors (VBTSs) have gained popularity in robotics systems. In this work, we introduce CrystalTac, a series of advanced optical tactile sensors with unique designs to each other, all fabricated through rapid monolithic manufacturing. Based on the previous work of C-Sight, we summarised the design and creation methods of known VBTS and proposed a new categorisation method to encapsulate their typical sensing mechanisms, including IMM, MDM, MFM, and multi-mechanism fusion. Then based on these typical mechanisms, five types of CrystalTatac sensors were fabricated using monolithical manufacturing. The experimental results showed that the newly designed Sensors effectively achieve their intended tactile sensing functions, offering assurance and motivation to other researchers working in the tactile robotics domain.

\end{abstract}

%%%%%% Main Text %%%%%%

\section{Introduction}
%, xu2022electrooculography

The integration of tactile sensors with robots has drawn considerable interest in fields ranging from soft robotics \cite{zhou2024integrated}, bionic robotics \cite{zhang2023bioinspired}, to human-robot interface areas \cite{haddadin2018tactile}. Among the different types of tactile sensors \cite{chi2018recent, wang2019tactile, liu2020recent}, vision-based tactile sensors (VBTSs) \cite{shimonomura2016robotic, shah2021design} use cameras to record the physical deformation when interacting with external objects. This method has gained increasing attention due to its superior spatial resolution on tactile sensing and relatively simple structure. The rapid adoption of VBTSs is largely due to advancements in computer vision. Vision sensors, such as RGB cameras, project a 3D scene into 2D frames, converting the colour, shape, and motion information of the external environment into a distribution of pixel values. In comparison, VBTSs operate similarly to reprographic devices, scanning and mapping 2.5D features \cite{johnson2009retrographic} into 2D images. These features, such as the texture geometry of the contact surface or shear force around the touched area, are termed 2.5D because their perception depth is confined to a limited range, and cannot be extended arbitrarily along the direction of light projection as 3D visual features can. Due to this limitation, VBTSs need a specific medium to convert physical information into optical signals that the camera can detect. The realisation of such a medium generally involves four steps: 
(1) establishing the correlation between tactile and optical features, often known as the tactile sensing mechanism; (2) creating the sensor architecture to embody this tactile sensing mechanism; (3) choosing appropriate techniques to manufacture the sensor's sub-components; and (4) finalizing the entire hardware assembly process based on the produced sub-components. In conclusion, the initial two steps fall under the `\textbf{design}' phase of VBTSs, while the final two steps comprise the `\textbf{creation}' phase. The design phase frequently imposes difficulties on the creation phase. This is due to the conventional method's segregation of design, manufacturing, and assembly stages. Nevertheless, in real-world applications, these stages are interconnected and must be meticulously coordinated \cite{zhang2022hardware}.

Among the three primary modules in VBTSs \cite{shah2021design}-illumination and vision modules-these are typically procured from commercial electronic component suppliers. In contrast, the contact module presents the most significant challenge \cite{fan2024magictac} (the subsequent discussion of VBTS manufacturing in this paper primarily refers to the fabrication of contact module). A VBTS contact module normally consists of six main components: the base, lens, elastomer, marker, coating, and skin. Although different types of VBTS may share similar core components, they are often produced using diverse manufacturing techniques \cite{fan2024magictac}. This multi-dimensional variation significantly influences the process from conceptual design to final assembly. As analysed in \cite{fan2024design}, the main challenges in VBTS manufacturing include complexities in design, process, time, and quality. Here, we briefly discuss how these four challenges impact the design and creation of VBTSs:

\textbf{i) Design}: VBTSs' structural framework and detailed constructions vary significantly, depending on the tactile sensing mechanism employed. This diversity presents a bidirectional challenge: (1) alterations in the sensor design may have a direct impact on the subsequent manufacturing stages, and (2) the constraints of existing manufacturing techniques can limit the initial sensor design.

\textbf{ii) Creation}: The creation of VBTSs is generally complex and involves several sub-components. The increase of manufacturing procedures necessitates more equipment, leading to substantial process and time complexities. Additionally, reliance on manual fabrication and assembly means the quality of the sensors depends on the specialised skills of the workforce. This results in a high incidence of assembly errors and variability in sensor performance.

Previous research has focused on the design and fabrication of a specific type of VBTSs \cite{zhang2022hardware, fan2024design, fan2024magictac, scharff2022rapid, li2022implementing}.  In \cite{fan2024design}, monolithic manufacturing technology was proposed to fabricate the C-Sight tactile sensor, which leverages advancements in multi-material 3D printing. By simplifying the VBTS manufacturing into a single printing sequence, this method accelerates production and ensures greater consistency and integration of the sub-components. It is expected to substantially improve the reliability, productivity, and affordability of the newly developed tactile sensor. However, previous work of C-Sight is limited to a single kind of sensing mechanism. To better explore the potential of this technology, we summarise the mainstream VBTSs into 5 categories based on their sensing principles and then propose to develop \textbf{CrystalTac-type sensor} following this categorisation, also known as \textbf{CrystalTac family}, aiming to demonstrate the versatility of monolithic manufacturing.

The name 'CrystalTac' draws on the properties of the crystal, a mineral known for its clarity and variability in colour and texture when combined with other minerals, just like the versatile structure of the different tactile sensing mechanism-based VBTSs. The monolithic manufacturing technology should be competent for the fabrication of the CrystalTac family by adapting to different design needs, including customised sensing principles, overall dimensions, and architectural details. 

The \textbf{main contributions} of this paper are listed as follows:

\begin{itemize}
\item We summarised the design and creation methods of known VBTSs and proposed a new categorisation method to encapsulate their typical sensing mechanisms, including IMM, MDM, MFM, and multi-mechanism fusion.

\item We developed the CrystalTac family, a series of sensors produced through monolithic manufacturing. This family includes C-Tac, C-Sight, C-SighTac, Vi-C-Tac, and Vi-C-Sight, each design based on a unique tactile sensing mechanism.

\item We conducted functional experiments to evaluate the sensing performance, cost-effectiveness, and design flexibility of CrystalTac-type sensors. Also, an optimisation upon sub-component manufacturing and several novel marker designs are introduced.

\end{itemize}

% \begin{figure*}[!htbp]
%     \centering
%     \includegraphics[width = 1.0\hsize]{figures/fig_1_v5_c.png}
%     \captionsetup{font=small} % Change 'small' to your desired size
%     \caption{A: Typical tactile sensing mechanism of known VBTSs. (a) IMM, such as GelSight and 9DTac\cite{yuan2017gelsight, lin20239dtact}; (b) MDM, such as TacTip and ChromaTouch\cite{ward2018tactip,scharff2022rapid}; (c,d,e) multi-mechanisms fusion, such as GelSlim, Finger Vision and TIRgel\cite{taylor2022gelslim, yamaguchi2016combining, zhang2023tirgel}. B: Monolithic manufacturing provides CrystalTac family superior design flexibility and creation efficiency. (a) An example of integrating various CrystalTac designs; (b) A real sample showcasing the integration of these designs through monolithic manufacturing.}
%     \label{VBTS_Principle}
%     \vspace{-0.1cm}
% \end{figure*}

\begin{figure*}[!htbp]
    \centering
    \includegraphics[width = 0.65\hsize]{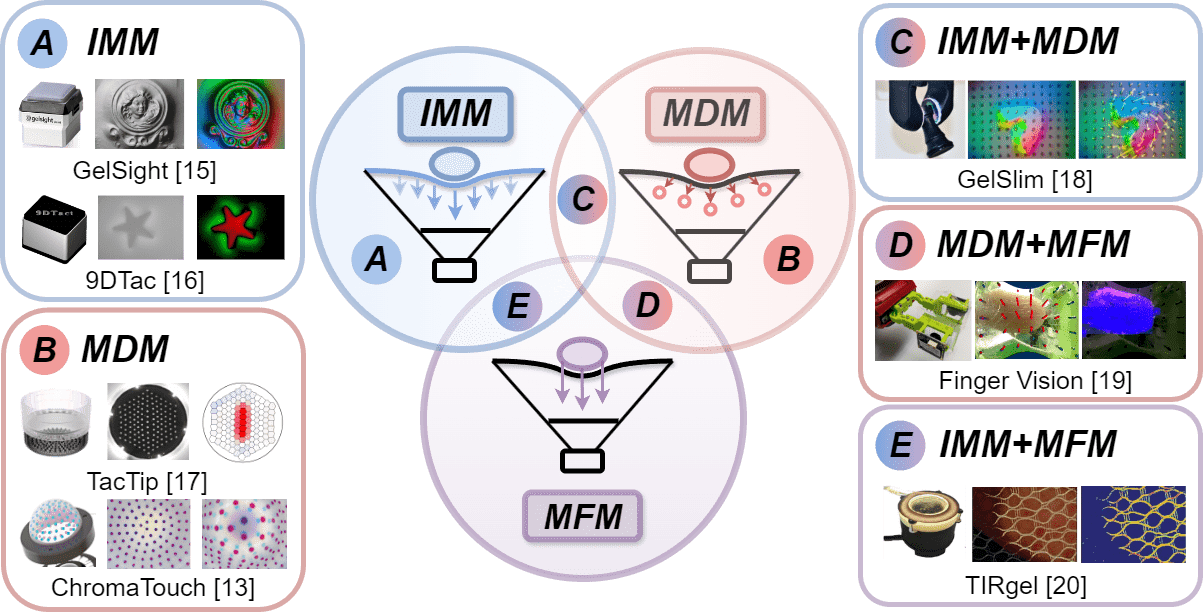}
    \captionsetup{font=small} % Change 'small' to your desired size
    \caption{Typical tactile sensing mechanism of known VBTSs. A: IMM, such as GelSight and 9DTac \cite{yuan2017gelsight, lin20239dtact}. B: MDM, such as TacTip and ChromaTouch \cite{ward2018tactip,scharff2022rapid}. C/D/E: multi-mechanism fusion consisting of IMM+MDM, MDM+MFM, and IMM+MFM, such as GelSlim, Finger Vision and TIRgel \cite{taylor2022gelslim, yamaguchi2016combining, zhang2023tirgel}.}
    \label{VBTS_Principle}
    \vspace{-0.3cm}
\end{figure*}

\section{Background}

Based on the four key steps for realising the `design' and `creation' of VBTSs introduced in the previous section, we review most of the known VBTSs and summarise their sensing mechanisms, manufacturing processes, and assembly methods.

\subsection{Typical Sensing Mechanism of Known VBTSs}

As shown in Fig. \ref{VBTS_Principle}, the sensing mechanisms of most VBTSs \cite{shah2021design, shimonomura2019tactile} can be categorised into several distinct methods: Intensity Mapping Method (IMM), Marker Displacement Method (MDM), Modality Fusion Method (MFM), and multi-mechanism fusion. 

\subsubsection{\textbf{IMM}}

The Intensity Mapping Method uses the pixel intensity values from the imaging frames to indicate tactile features, as illustrated in Fig. \ref{VBTS_Principle}(A).
This approach leverages the high resolution of the VBTS camera frame and the changes in each pixel value over a threshold interval to reconstruct continuous features, including fine geometric textures and dense motion distributions. This change in pixel value requires the VBTS to incorporate specific structures, such as a light-conductive plate or a reflective coating. For example, the earlier optical waveguide-type sensors \cite{maekawa1992development,ikai2016robot, hiraishi1988object, shimonomura2016robotic} usually rely on the total internal reflective (TIR) \cite{mossman2004novel} to map the contact pattern located on the light-conductive plate surface. GelSight-type sensors \cite{yuan2017gelsight}, such as OmniTact \cite{padmanabha2020omnitact}, Digit \cite{lambeta2020digit}, GelTip \cite{gomes2020geltip}, InSight \cite{sun2022soft}, and Tactile Fingertip \cite{romero2020soft}, employ a coating layer with uniform reflective properties applied to a silicone elastomer, enabling pixel-level tactile sensing. By incorporating RGB lighting, GelSight captures multiple images under different lighting conditions within a single frame, facilitating the estimation of the overconstrained gradient of the contact surface geometry through the photometric stereo algorithm \cite{woodham1980photometric, johnson2009retrographic}. In contrast to intensity mapping via a light-conductive plate or a reflective coating, DTac-type sensors \cite{lin2023dtact}, such as 9DTac \cite{lin20239dtact}, employ a combination of translucent gel and an opaque layer, where variations in pixel darkness can indicate intensity-to-depth regressions or even estimate force/torque. Furthermore, if a depth camera is used instead of a traditional RGB camera in VBTS, each pixel intensity value will also include real physical distance information in addition to colour information. \cite{li2021design} integrates a latex inflatable film with a depth camera, achieving precise deformation measurements through stereo vision.

\subsubsection{\textbf{MDM}}
The Marker Displacement Method, depicted in Fig. \ref{VBTS_Principle}(B), translates tactile information into displacement distributions of marker patterns either positioned on the surface or embedded within the elastomer. In \cite{li2023marker}, MDM is classified into 2D-MDM, 2.5D-MDM, and 3D-MDM, based on the type of tactile features represented by the pattern changes. Both 2D-MDM and 2.5D-MDM employ a single camera to capture the two-dimensional movements of markers in horizontal and vertical directions. However, 2.5D-MDM can indirectly represent corresponding contact depth information by observing the deformable characteristics of the markers, such as changes in size or shape. In contrast, 3D-MDM precisely captures the depth of markers using stereo vision. GelForce \cite{vlack2005gelforce, sato2009finger} uses two layers of spherical marker arrays to measure 3D force distributions. Inspired by the dermal papillae in the human fingertip, TacTip \cite{lepora2021soft, ward2018tactip, lepora2022digitac} designs pin-shaped markers, which amplify contact information through leverage principles. Furthermore, Yang et al.\cite{yang2021enhanced} employ marker displacements to estimate the gradient of the contact surface through the integration method. ChromaTouch \cite{scharff2022rapid} introduces two superimposed colour filters as a composite marker pattern, where contact deformation alters the markers' hue, centroid, and apparent size. Similar designs are found in \cite{lin2019sensing, lin2020curvature}. DelTact \cite{zhang2022deltact}, Viko \cite{pang2021viko}, and \cite{du2021high} all utilise a dense random colour pattern for precise tracking of contact deformations. Sferrazza et al.\cite{sferrazza2019design} randomly embed spherical fluorescent green markers within an elastomer to enhance the sensing range. A multi-colour continuous marker pattern (CMP) \cite{li2023improving} has been developed to improve the representation and extraction of VBTS contact information. Tac3D \cite{zhang2022tac3d} employs virtual binocular vision to precisely measure the three-dimensional shape and force distribution of the contact surface using CMP.

\subsubsection{\textbf{MFM}}
The Modality Fusion Method integrates multi-modal features beyond sole tactile input, typically incorporating vision and proximity sensing. This integration effectively mitigates the inherent limitations of a single tactile modality, such as the inability to perceive features like colour and distance. However, MFM cannot operate independently in VBTSs and needs to be combined with IMM or MDM for tactile sensing, thereby included in the concept of multi-mechanism fusion.

\subsubsection{\textbf{Multi-Mechanism Fusion}}
The sensing mechanisms for VBTS introduced above each offer distinct advantages. However, researchers have been proactively developing multi-mechanism fusion sensors to further improve the performance of VBTSs by obtaining comprehensive sensing information.

\begin{itemize}

\item \textbf{IMM + MDM}: GelSight-type sensors \cite{yuan2017gelsight} incorporate dot markers above the reflective coating layer, aiming for force and shear detection rather than solely capturing fine textures, as demonstrated by GelSlim \cite{taylor2022gelslim}, shown in Fig. \ref{VBTS_Principle}(C), GelSight Wedge \cite{wang2021gelsight}, and GelStereo \cite{zhang2023gelstereo}. DenseTact \cite{do2023densetact} employs a randomised dense pattern with a coating to extract continuous tactile features. The Soft-bubble sensor\cite{kuppuswamy2020soft} utilises a stereo camera for the tracking of shear-induced displacement through custom marker patterns. \cite{wang2021novel} introduced Particle Image Velocimetry (PIV) to establish a linear relationship between force and velocity. UVtac \cite{kim2022uvtac} employs a switchable ultraviolet (UV) method to decouple marker patterns and reflective membrane images, facilitating both object localisation and force estimation, with each functionality remaining independent of the other.

\item \textbf{MDM + MFM}: Finger Vision \cite{yamaguchi2016combining}, as illustrated in Fig. \ref{VBTS_Principle}(D), embeds black marker patterns on a transparent elastomer to achieve multi-modality sensing. The approach proposed by \cite{yamaguchi2021fingervision} increases force resolution and sensitivity of Finger Vision by introducing whiskers. ViTacTip \cite{fan2024vitactip} also employs a transparent skin to enhance TacTip \cite{ward2018tactip} with vision-tactile fusion capability. Similar to UVtac, SpecTac \cite{wang2022spectac} uses UV LEDs and randomly distributed UV fluorescent markers, allowing for a switch between visual and tactile sensing modes, controlled by the activation of the UV LEDs.

\item \textbf{IMM + MFM}: Shimonomura et al.\cite{shimonomura2016robotic} employ a light-conductive plate and compound-eye camera to capture both tactile and proximity modality information. The tactile feature is obtained as an infrared image through the TIR principle, while the proximity feature is detected through stereo matching of a pair of images obtained by two cameras. See-Through-your-Skin (STS) \cite{hogan2021seeing} utilises a translucent coating and adjustable lighting to enable the transition between tactile and visual sensing modalities. Compared to STS, StereoTac \cite{roberge2023stereotac} and VisTac \cite{athar2023vistac} incorporate binocular cameras, extending 2D visual sensing into 3D areas. As displayed in Fig. \ref{VBTS_Principle}(E), TIRgel \cite{zhang2023tirgel} also utilises TIR through a purely transparent elastomer to visualise tactile features, achieving conversion between visual and tactile modalities via focus adjustment.

\end{itemize}

\renewcommand{\arraystretch}{1.5} % 增加行高倍数以增加行间距

\begin{table}[!htbp]
\caption{Comparison of Typical VBTSs on Tactile Sensing Mechanism and Manufacturing Method}
\resizebox{\textwidth}{35mm}{
\begin{tabular}{cccccccc}
\hline

\textbf{Sensor}         & \textbf{Mechanism} & \textbf{Base} & \textbf{Lens}      & \textbf{Elastomer} & \textbf{Marker} & \textbf{Skin} & \textbf{Coating} \\ 
\hline
\textit{\textbf{Digit}} \cite{lambeta2020digit}         & IMM                        & Mould-formed  & Laser-cut          & Mould-formed       & -               & -             & Airbrushed       \\
\textit{\textbf{OmniTact}} \cite{padmanabha2020omnitact}       & IMM                        & 3D-printed    & -                  & Mould-formed       & -               & -             & Hand-painted     \\
\textit{\textbf{9DTac}} \cite{lin20239dtact}         & IMM                        & 3D-printed    & Laser-cut          & Mould-formed       & -               & Mould-formed  & -                \\ 
\textit{\textbf{*}} \cite{zhang2023novel}         & IMM                        & 3D-printed    & Laser-cut          & DIY-modified       & -               & -  & Hand-gilded                \\ \hline
\textit{\textbf{TacTip}} \cite{lepora2021soft, ward2018tactip, lepora2022digitac}          & MDM                        & \textbf{3D-printed}    & Laser-cut          & Injection-filled   & \textbf{3D-printed}      & \textbf{3D-printed}    & -                \\
\textit{\textbf{ChromaTouch}} \cite{scharff2022rapid}    & MDM                        & \textbf{3D-printed}    & -                  & Injection-filled   & \textbf{3D-printed}      & \textbf{3D*} + Mould*    & -                \\
\textit{\textbf{DelTact}} \cite{zhang2022deltact}        & MDM                        & 3D-printed    & Laser-cut          & Mould-formed       & Film-sticked    & Airbrushed    & -                \\ \hline
\textit{\textbf{GelSlim3.0}} \cite{taylor2022gelslim}     & IMM+MDM                    & 3D-printed    & Commercial-ordered & Mould-formed       & Ink-printed     & Airbrushed    & Airbrushed       \\
\textit{\textbf{GelSight Wedge}} \cite{wang2021gelsight} & IMM+MDM                    & 3D-printed    & Laser-cut          & Mould-formed       & Ink-printed     & Film-sticked  & Airbrushed       \\
\textit{\textbf{DenseTac2.0}} \cite{do2023densetact}   & IMM+MDM                    & 3D-printed    & Laser-cut          & Mould-formed       & Ink-printed     & -             & Paint-dipping    \\
\textit{\textbf{UVtac}} \cite{kim2022uvtac}          & IMM+MDM                    & 3D-printed    & Laser-cut          & Mould-formed       & Ink-printed     & -             & Airbrushed       \\ \hline
\textit{\textbf{Finger Vision}} \cite{yamaguchi2021fingervision}  & MDM+MFM                    & 3D-printed    & Laser-cut          & Mould-formed       & Solid-embedded  & Film-sticked  & -                \\
\textit{\textbf{ViTacTip}} \cite{fan2024vitactip}  & MDM+MFM                    & \textbf{3D-printed}    & Laser-cut          & Injected-filled       & \textbf{3D-printed}  & \textbf{3D-printed}  & -                \\
\textit{\textbf{SpecTac}} \cite{wang2022spectac}        & MDM+MFM                    & 3D-printed    & Laser-cut          & Mould-formed       & Hand-painted    & Hand-painted  & -                \\ \hline
\textit{\textbf{STS}} \cite{hogan2021seeing}            & IMM+MFM                    & 3D-printed    & Laser-cut          & Mould-formed       & -               & Airbrushed    & Airbrushed       \\
\textit{\textbf{VisTac}} \cite{athar2023vistac}         & IMM+MFM                    & 3D-printed    & Laser-cut          & Mould-formed       & -               & Hand-painted  & Airbrushed       \\
\textit{\textbf{TIRgel}} \cite{zhang2023tirgel}         & IMM+MFM                    & 3D-printed    & Laser-cut          & Mould-formed       & -               & -             & -                \\ 
\hline
\end{tabular}}
\label{VBTS manufacturing compare}
% \vspace{-0.5cm}
\end{table}

\subsection{Typical Manufacturing Method of Known VBTSs}

Based on Table \ref{VBTS manufacturing compare}, the manufacturing methods of known VBTSs's sub-components are discussed below. The different tactile sensing mechanisms lead to variations in their manufacturing processes, where multi-mechanism fusion further complicates the procedure.

\subsubsection{\textbf{Base}}

The base, also known as the case, mount, frame, body, bracket, or housing, typically serves as the external structure of VBTSs. It acts as the connecting element between the contact module and the illumination/vision modules. To ensure a secure fit among these modules and to achieve the required durability and compactness of the sensors, the base must be designed with an appropriate shape and constructed from a material of sufficient rigidity. 3D printing technology \cite{yuan2017gelsight,lin20239dtact,pang2021viko, padmanabha2020omnitact, yang2021enhanced, du2021high, zhang2023tirgel, do2023densetact, kuppuswamy2020soft, romero2020soft, kim2022uvtac, li2024biotactip} is the most common option for prototyping. While commercially large-scale productions usually rely on the mould-forming method \cite{lambeta2020digit}.

\subsubsection{\textbf{Lens}}

The lens allows the camera to capture tactile information without obstruction. Additionally, it internally supports the elastomer during interactions. The common manufacturing method employs a laser cutter to shape the acrylic board \cite{yuan2017gelsight, athar2023vistac, lin2023dtact, pang2021viko, sato2009finger, zhang2022deltact, li2024biotactip, zhang2023tirgel, yamaguchi2016combining, kuppuswamy2020soft, roberge2023stereotac}. However, lenses with complex curved surfaces, diverging from simple planar shapes, can only be produced through mould-forming methods \cite{romero2020soft} or procured from commercial suppliers \cite{taylor2022gelslim}. Additionally, some sensors \cite{scharff2022rapid, sun2022soft, padmanabha2020omnitact} do not incorporate lenses but rely on their inherent structure or an internal skeleton.

\subsubsection{\textbf{Elastomer}}

The elastomer serves as one of the primary mediums for converting tactile information into visual data, with core properties including transparency, colour, and stiffness. Mould-formed silicone \cite{abad2020visuotactile, zhang2022hardware} is the material most preferred for creating these elastomers, as it offers adjustable properties to suit various applications \cite{athar2023vistac,yuan2017gelsight, pang2021viko, taylor2022gelslim, yang2021enhanced, zhang2022deltact, du2021high, sferrazza2019design, zhang2023tirgel, yamaguchi2016combining, do2023densetact, roberge2023stereotac, hogan2021seeing, kim2022uvtac, scharff2022rapid, li2022implementing}. The preparation process involves  A/B solutions mixing, mould casting, bubble removal, and heat curing. Translucent and coloured silicone may require dyeing with pigment \cite{lin20239dtact}. The adjustable stiffness depends on the variation of the ratio of mixed solutions \cite{romero2020soft}. Some VBTSs use commercial products as alternatives, modified using DIY methods \cite{abad2020low, zhang2023novel}, to simplify manufacturing. There are exceptions, such as Soft-bubble \cite{kuppuswamy2020soft}, which features an air-filled membrane design, while TacTip \cite{lepora2021soft}, BioTacTip \cite{li2024biotactip} and \cite{ito2012vision} use mixed ultra-soft gel (TechsiL RTA27905 A/B) and coloured water respectively.

\subsubsection{\textbf{Marker}}

For MDM-type VBTSs, markers play the role of mapping tactile deformation to visual pattern distribution. Inspired by \cite{zhang2022hardware}, we classify the marker manufacturing methods into three types:

\begin{itemize}
\item \textbf{Surface Fabrication}: Markers are prepared on the surface of premade elastomers through direct or indirect methods. The former approach combines light etching, mask templates, and stamp plates with ink printing \cite{taylor2022gelslim, pang2022viko, wang2021novel, kim2022uvtac, do2023densetact}, or utilises plastic beads \cite{yamaguchi2016combining, yang2021enhanced}. Similarly, SpecTac \cite{wang2022spectac} and \cite{abad2020low} manually apply fluorescent markers using a brush and UV pen, respectively. In the latter case, the complete marker patterns are printed in advance using materials such as sticker film \cite{du2021high}, transfer paper film \cite{zhang2023gelstereo}, or adhesive-backed templates \cite{kuppuswamy2020soft}, and then applied to the surface of the elastomer.

\item \textbf{Embedded Fabrication}: To capture the deformation field at different depths, rather than solely at the contact plane, markers are embedded within the elastomer. \cite{sferrazza2019design} incorporated fluorescent markers into the solutions during the preparation of the silicone. GelForce \cite{vlack2005gelforce, sato2009finger} and \cite{lin2019sensing, lin2020curvature} placed two marker arrays of different colours in elastomer, prepared layer by layer.

\item \textbf{Integral Fabrication}: 

Markers are prepared integrally with the skin in a single piece. The pin-shaped markers with the skin of TacTip were initially produced by mould casting \cite{winstone2012tactip}. Subsequently, 3D printing has been introduced for the later versions of  MDM-type VBTSs \cite{lepora2021soft, lepora2022digitac, fan2024vitactip, li2024biotactip}.

\end{itemize}

\subsubsection{\textbf{Skin}}

The skin serves as the direct contact interface with the external environment. For VBTSs designed with silicone elastomer as the main body, some lack a skin layer \cite{yuan2017gelsight, yamaguchi2016combining, padmanabha2020omnitact, zhang2022deltact, zhang2023tirgel, zhang2023novel, kim2022uvtac}, while others employ a skin for protection, which is achieved by casting a thin layer of silicone through mould-forming \cite{athar2023vistac, yang2021enhanced} or spray painting \cite{roberge2023stereotac, kim2022uvtac, hogan2021seeing}. To filter noise, DTac \cite{lin2023dtact, lin20239dtact} manually applies black silicone on the top surface, similar to \cite{sato2009finger, du2021high, sferrazza2019design}. Several designs implement fabric films or adhesive tape films as a skin-like layer, as demonstrated in \cite{zhang2022tac3d, ma2019dense, wang2021gelsight}. For example, Finger Vision \cite{yamaguchi2016combining} employs a thin transparent plastic film to protect the silicone body from contaminants. Some VBTSs rely on the skin as the primary structural component. TacTip \cite{lepora2021soft, ward2018tactip, lepora2022digitac} and ChromaTouch \cite{scharff2022rapid} build the skin as the main body, with the latter also requiring the casting of a white silicone layer to filter noise. Furthermore, soft bubbles \cite{kuppuswamy2020soft} and \cite{li2021design} both use hand-cut latex film as an outer skin for an air-filled structure.

\subsubsection{\textbf{Coating}} \label{manufacturing coating}

Coating layers are widely used in IMM-VBTSs. Unlike the skin, which is primarily intended for protection and shading, coatings are designed to convert tactile features into visual imaging. These can be categorised into two broad types: reflective coatings and controllable coatings.

\begin{itemize}
\item \textbf{Reflective Coating}: GelSight-type sensors \cite{yuan2017gelsight} incorporate metal pigments into silicone, using bronze flakes for semi-specular coatings and fine aluminum powder for matte coatings. Various manufacturing methods are employed, such as brush painting \cite{padmanabha2020omnitact}, airbrushing \cite{abad2020low, lambeta2020digit, taylor2022gelslim}, sputtering \cite{jiang20183}, and the paint dipping technique \cite{do2023densetact}. Similarly, \cite{zhang2023novel} uses metal foil to create a semi-mirror coating through the gilding process.

\item \textbf{Controllable Coating}: Some MFM-based VBTSs control the coating transparency by adjusting internal lighting conditions. STS \cite{hogan2021seeing}, StereoTac \cite{roberge2023stereotac}, and VisTac \cite{athar2023vistac} all apply 2-3 layers of `mirror spray' to achieve a translucent layer as the controllable coating.

\end{itemize}

\subsection{Typical Assembly Method of Known VBTSs}

The VBTS assembly strategy is often overlooked in most of the current research. Here, the assembly process is analyzed across three dimensions: tools, workflows, and mechanisms.

\subsubsection{\textbf{Assembly Tool}} \label{Assembly Tool}

The assembly tool consists of \textbf{manual assembly} and \textbf{machine assembly}. Due to the widespread use of traditional manufacturing methods, such as mould-forming \cite{yuan2017gelsight, yamaguchi2021fingervision}, manual assembly remains the prevalent practice. Machine assembly, characterised by the use of specific devices such as multi-material 3D printers, is exemplified by TacTip \cite{ward2018tactip, lepora2022digitac} and ChromaTouch \cite{scharff2022rapid}. Generally, manual assembly may result in assembly errors, while machine assembly necessitates specific hardware equipment. Most VBTSs are assembled manually or in hybrid form, with a limited number being fully assembled by machines, such as MagicTac \cite{fan2024magictac}.

\subsubsection{\textbf{Assembly Workflow}} \label{Assembly Workflow}

The assembly workflow consists of \textbf{serial assembly} and \textbf{parallel assembly}. Serial assembly involves combining components in a specific sequential order, typically determined by structural design or manufacturing requirements. For example, 9DTac \cite{lin20239dtact} consists of a transparent layer, a translucent layer, and a black layer, thereby asking for a fixed assembly order due to its stacked construction. In contrast, parallel assembly allows components to be assembled simultaneously, as seen in \cite{shimonomura2016robotic}, which achieves this due to its simple structure. Although parallel assembly is more efficient, it presents challenges in the management of sensors requiring complex processes and intricate structures. The majority of VBTSs are fabricated using serial assembly, while a minority employ a hybrid of serial and parallel assembly. Purely parallel assembly is rarely used.

\subsubsection{\textbf{Assembly Mechanism}} \label{Assembly Mechanism}

The assembly mechanism consists of \textbf{physical assembly} and \textbf{chemical assembly}. Physical assembly relies solely on the interaction of hardware structures. For instance, \cite{sferrazza2019design} involves casting silicone into the base mould, thereby directly assembling the elastomer with the base. Chemical assembly utilises the properties of chemical reagents, such as material compatibility and adhesion. For example, coating pigments are often mixed with silicone to enhance adhesion with the elastomer, and lenses are frequently attached to the base using adhesives. Although physical assembly offers the benefits of convenience and is well-suited for modular designs, chemical assembly excels in ensuring enhanced durability and stability. Most VBTSs are produced through a hybrid of physical and chemical assembly, with only a few being made entirely through physical or chemical assembly.

\begin{figure*}[!htbp]
    \centering
    \includegraphics[width = 1.0\hsize]{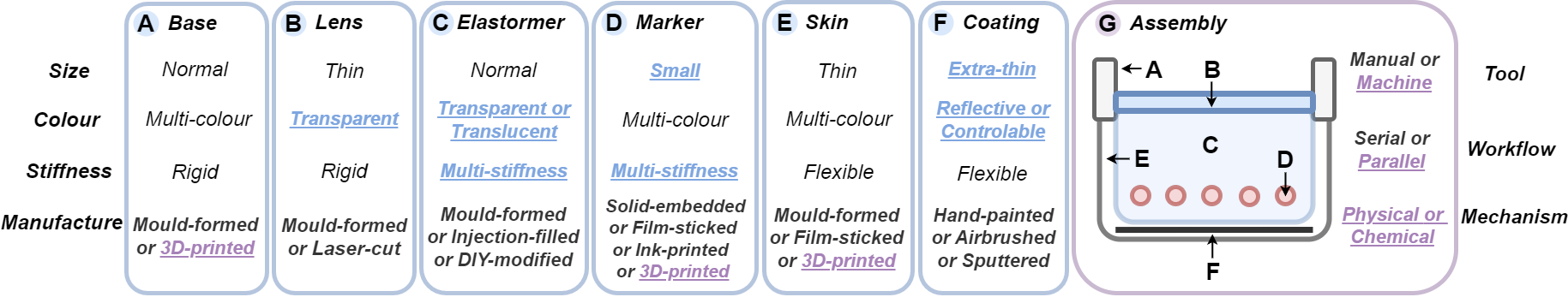}
    \captionsetup{font=small} % Change 'small' to your desired size
    \caption{Representative attributes of VBTS. A(Base): the external supporting structure; B(Lens): the transparent medium for camera imaging; C(Elastormer): the flexible main body; D(Marker): the physical medium for visualizing tactile information; E(Skin): the external layer for shading or protection; F(Coating): the functional layer for contact feature mapping; G(Assembly): the method of installing the aforementioned sub-components, involving various tools, workflows, and mechanisms. 
    }
    \label{VBTS_attribute}
    % \vspace{-0.6cm}
\end{figure*}

In this section, we review the commonly used methods in the design and creation of VBTS, whose representative attributes can be summarised in Fig. \ref{VBTS_attribute}. Each VBTS has several attributes with distinct characteristics, such as size, colour, stiffness, and manufacturing methods. This complexity corresponds to a wide range of structural variations, placing high demands on subsequent manufacturing and assembly processes. The complexities in process, time, and quality are difficult to address due to the limitations of traditional manufacturing methods. In \cite{fan2024design}, a rapid monolithic manufacturing technique has been proposed which has the potential to become a standard method to simplify the design and creation of VBTS. Its feasibility has been evaluated by C-Sight, which relies on IMM. In this work, the CrystalTac family is designed as a series of sensors with different sensing mechanisms to demonstrate the adaptability of monolithic manufacturing.

\section{Design and Creation}\label{design and fabrication}

Based on the details of monolithic manufacturing elaborated in \cite{fan2024design}, this work focuses on the design and creation of the CrystalTac family. The first step involves discussing the manufacturing feasibility of CrystalTac-type sensors based on different sensing mechanisms through monolithic manufacturing, followed by the overall design of the CrystalTac family.

\subsection{Manufacturing Feasibility of Different Sensing Mechanism-based CrystalTac}

From Table \ref{VBTS manufacturing compare} and Fig. \ref{VBTS_attribute}, it is evident that different sensing mechanism-based VBTSs exhibit certain tendencies in terms of sub-components, each associated with specific attributes and corresponding manufacturing methods. In detail, the choice of sensing mechanism directly determines the type of sub-component, while the specific sensor design based on that mechanism influences the attributes of each sub-component. The most striking issue is the significant variation in size, colour, and stiffness requirements for the various sub-components. Here, we discuss which sensing mechanism could be applied to CrystalTac-type sensors in terms of technological and practical feasibility.

\subsubsection{Technological Feasibility}

As utilised in monolithic manufacturing, PolyJet Printing (PP) \cite{muthuram2022review} functions similarly to inkjet printing by spraying thousands of photopolymer droplets rather than ink. This process employs ultraviolet (UV) light to cure and construct parts in a layer-by-layer fashion. The combination of inkjet and photopolymerization technologies provides PP with two significant advantages:

\begin{itemize}

\item \textbf{High Printing Quality}: Due to the small size of the ejected resin droplets, impressive micron-level spatial resolution can be achieved in both the horizontal XY-direction and vertical Z-direction. This level of precision is essential for achieving high print quality, characterised by fine resolution and a superior surface finish.

\item \textbf{Multi-material Printing}: By integrating multiple print heads, pp easily achieves multi-material printing with consistent print quality, whose printing capabilities encompass a wide variety of material properties. In addition to multi-colour printing, it can also produce flexible materials with varying properties.

\end{itemize}

\begin{figure}[!htbp]
    \centering
    \includegraphics[width = 1.0\hsize]{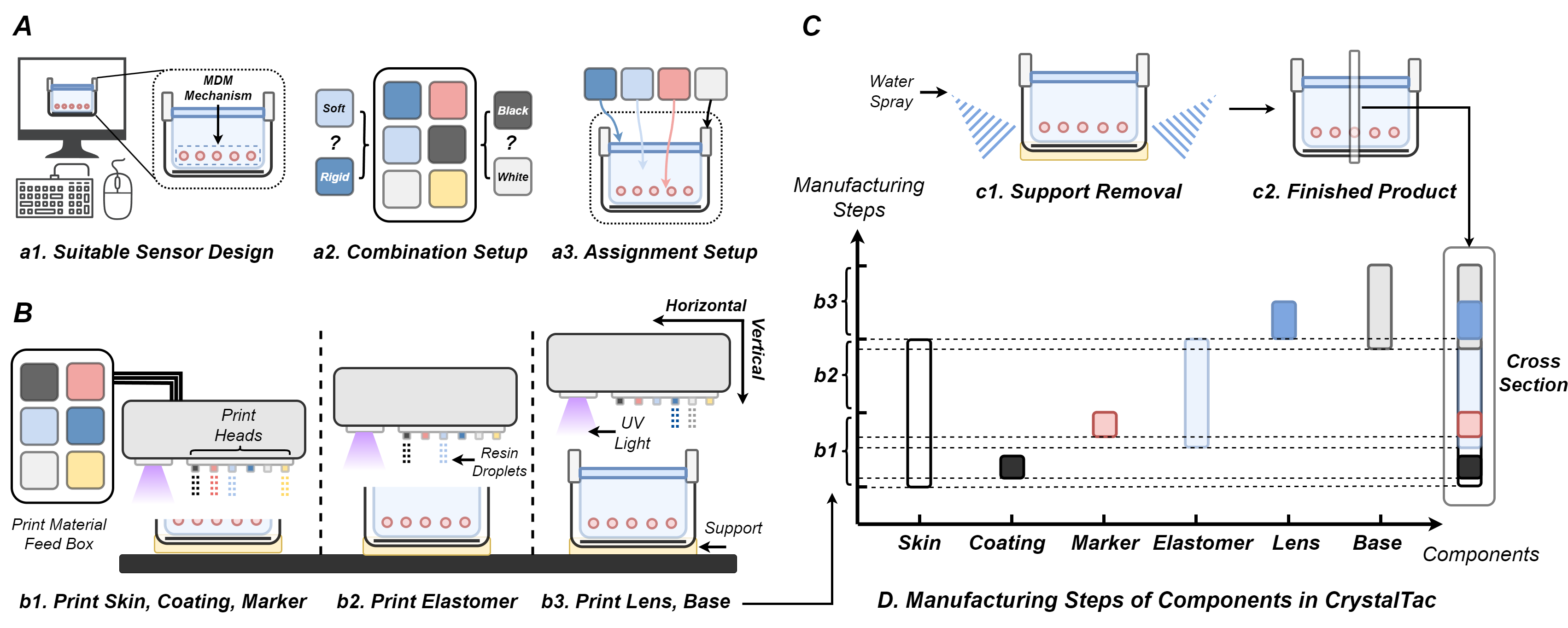}
    \captionsetup{font=small} % Change 'small' to your desired size
    \caption{A: The practical feasibility of CrystalTac depends on the suitable sensor design and the setup limitations of the print materials. B: Monolithic manufacturing integrates component fabrication and assembly into a single process. C: The finished CrystalTac is ready for use after removing support structures, either through water spray or manual tools. D: CrystalTac can realise complicated structural designs.}
    \label{VBTS_PP}
    % \vspace{-0.6cm}
\end{figure}

These two advantages have considerable significance in manufacturing different sensing mechanism-based CrystalTac with complicated structure, as shown in Fig. \ref{VBTS_PP}(B-D). Referring to the typical print materials listed in \cite{fan2024design}, Vero series (VB/VW/VC)\footnote{https://www.stratasys.com/en/materials/materials-catalog/polyjet-materials/vero/}, Agilus30 series (AB/AW/AC)\footnote{https://www.stratasys.com/en/materials/materials-catalog/polyjet-materials/agilus30/}, and support materials\footnote{https://www.stratasys.com/en/materials/materials-catalog/polyjet-materials/polyjet-support-materials/}, are suitable selections for the skin, marker, elastomer, lens, and base of CrystalTac. However, due to the lack of print materials containing metal powder, reflective or controllable coatings are currently unavailable, rendering some IMM-VBTSs based on these coatings, such as GelSight, not yet feasible for CrystalTac.

\subsubsection{Practical Feasibility}

As displayed in Fig. \ref{VBTS_PP}(A), the practical feasibility depends on the suitable design of each sensor, as well as the combination and assignment setups of loaded print materials.

\begin{itemize}

\item \textbf{Suitable Sensor Design}: As illustrated in Fig. \ref{VBTS_PP}(A.a1), the structural design of CrystalTac must be based on a defined mechanism. For example, MDM-CrystalTac requires the marker pattern to be embedded in the correct position, MFM-CrystalTac often requires minimal occlusion within the imaging path, and IMM-CrystalTac may need a multi-layer stacked structure. An irrational structural design can fail to achieve the desired sensing mechanism. However, the efficiency and flexibility of monolithic manufacturing can help researchers rapidly implement hardware iterations of a prototype design by controlling variables.

\item \textbf{Combination Setup of Print Material}: It should be noted that the print materials loaded in the feed box are limited, and the number of print heads is also fixed by the printer version, as shown in Fig. \ref{VBTS_PP}(A.a2). Therefore, it is necessary to consider the actual material loading situation of the current printer, especially with multi-mechanism fusion designs which require a more complex range of print materials. In general, the minimum criteria to be fulfilled include at least three print materials, of which AC, VW/VB, and support materials must be loaded. If there are spare print heads available, priority should be given to loading AW/AB or VC. Such strategies can maximise the design options for CrystalTac with limited material combinations.

\item \textbf{Assignment Setup of Print Material}: After determining the sensor design, it is necessary to assign the appropriate material to each sub-component based on different attributes as shown in Fig. \ref{VBTS_PP}(A.a3), considering various factors such as colour, stiffness, and shape. For example, MFM-CrystalTac requires high transparency and low distortion levels in the skin, while MDM-CrystalTac requires effective shading. It should be emphasised that although the number of print materials loaded is limited, their properties can be adjusted through fusion with each other. For instance, by mixing flexible Agilus30 materials into the rigid Vero series, materials with intermediate stiffness can be obtained, while Vero Vivid\footnote{https://www.stratasys.com/en/materials/materials-catalog/polyjet-materials/verovivid/} can be combined with a wide range of materials to achieve rich colours. These characteristics can be adjusted by changing the mixing ratio, and the entire process is automated by the printer, greatly enhancing design flexibility and ensuring production quality.

\end{itemize}

The above analysis demonstrates that monolithic manufacturing can provide sufficient capability to create the CrystalTac family with different sensing mechanisms but given certain prerequisites. As seen in \textbf{Supplementary \ref{Supplementary 5.1}}, taking into account the optimisation for key sub-components of CrystalTac, a thickness range of 2-3 mm is optimal for printed lenses, offering a balance between high structural strength and satisfactory imaging performance. Similarly, the suitable height for the printed elastomer should range between 2 and 5 mm to balance image quality and softness.

\subsection{Overal Design of CrystalTac Family}

The CrystalTac family design is introduced here, whose family tree is illustrated in Fig. \ref{crystal_family_tree}. According toFig. \ref{VBTS_Principle}(A), the CrystalTac family is categorised into five branches, the C-Sight using IMM mechanism, the C-Tac using MDM mechanism, the C-SighTac using IMM + MDM mechanism, the Vi-C-Sight using IMM + MFM mechanism and the Vi-C-Tac using MDM + MFM mechanism. It should be emphasised that these five CrystalTac sensors are proposed not to provide a final design version, but rather to validate its potential to implement various tactile sensing mechanisms.

\subsubsection{\textbf{Contact Module}}

\begin{itemize}

\item \textbf{C-Sight}: To realise the IMM mechanism, C-Sight has been designed, as described in \cite{fan2024design}, inspired by DTac \cite{lin2023dtact, lin20239dtact}. The sectional diagram in Fig. \ref{crystal_family_tree}(A) shows that within the gap between C-Sight's outer skin (a6) and the clear elastomer (a3), two additional components are present: a translucent layer (a5) composed of support material and a pure white layer (a4) made of Agilus30 White. The deformation resulting from external contact alters the distance between the black skin and the white layer. Where the distance is shortened, the pixel intensity appears darker, aiding in inferring the contact depth.

\item \textbf{C-Tac}: To achieve the MDM mechanism, C-Tac can be embedded with 2D and 2.5D markers through different material assignments as shown in Fig. \ref{crystal_family_tree}(B), such as rigid Vero and flexible Agilus30 series. The distinction between them lies in the ability to provide pseudo-depth information, achieved either through shape deformation or variations in morphological structure. Monolithic manufacturing allows for customisation of both the stiffness properties and the morphological geometry of the markers. As shown in \textbf{Supplementary \ref{Supplementary 5.2}}, three kinds of marker patterns can be employed for C-Tac, including multi-layer markers, Voronoi markers and coordinate markers, each possessing distinct characteristics compared to dot markers.

\item \textbf{C-SighTac}: As illustrated in Fig. \ref{crystal_family_tree}(C), C-SighTac is developed based on C-Sight and C-Tac to achieve the IMM+MDM mechanism. This approach involves the structural framework of C-Sight and marker designs from C-Tac into appropriate positions. Similar to marker-enhanced GelSight \cite{taylor2022gelslim}, markers may enable C-SighTac more sensitive to dynamic features, such as force.

\item \textbf{Vi-C-Tac}: Inspired by Finger Vision \cite{yamaguchi2021fingervision} and ViTacTip \cite{fan2024vitactip}, Vi-C-Tac uses transparent skin to replace the opaque one of C-Tac as displayed in Fig. \ref{crystal_family_tree}(d), aiming to realise MDM+MFM mechanism. The internally embedded markers provide greater dynamic sensitivity to physical interaction while retaining the multi-modality sensing capability through the clear skin.

\begin{figure*}[!htbp]
    \centering
    \includegraphics[width = 0.95\hsize]{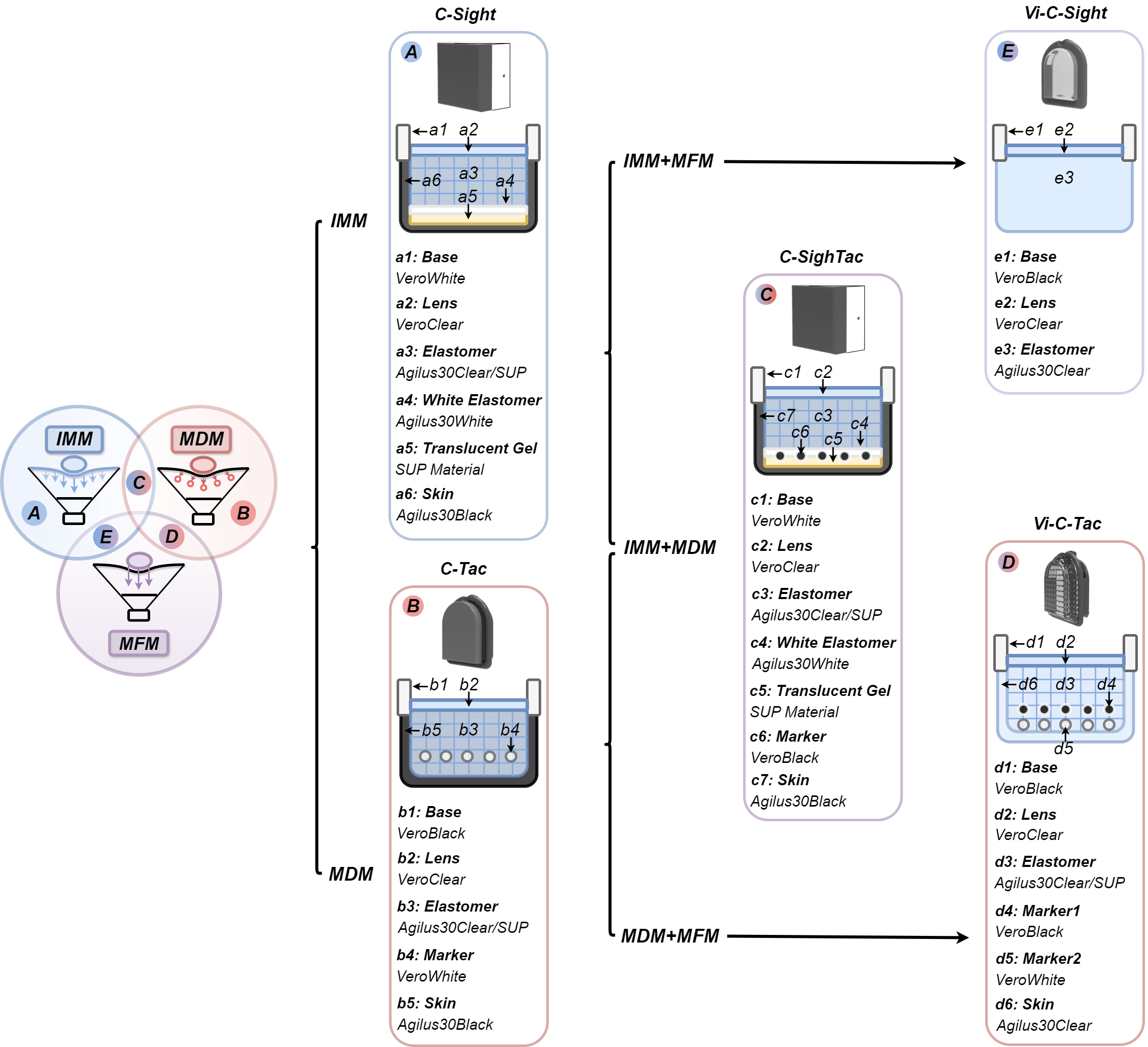}
    \captionsetup{font=small} % Change 'small' to your desired size
    \caption{The family tree of CrystalTac comprises five major branches, each proposed with a different tactile sensing mechanism. A: C-Sight, IMM mechanism. B: C-Tac, MDM mechanism. C: C-SighTac, IMM+MDM mechanism. D: Vi-C-Tac, MDM+MFM mechanism. E: Vi-C-Sight, IMM+MFM mechanism. The diagram for each design illustrates its unique internal structure and material assignment.}
    \label{crystal_family_tree}
\end{figure*}

\item \textbf{Vi-C-Sight}: Vi-C-Sight utilises IMM+MFM mechanism, inspired by \cite{shimonomura2016robotic}. Using TIR principle, variations in the gradient of the elastomer surface upon the contacted object alter the trajectory of internally reflected light beams. This causes the contact texture to be mapped onto the transparent elastomer surface, resulting in imaging brightness change while still retaining background visual information. As seen in Fig. \ref{crystal_family_tree}(E), Vi-C-Sight uses pure Agilus30 Clear as the elastomer material to leverage this mechanism rather than a multi-layer grid structure.

\end{itemize}

\subsubsection{\textbf{Vision and Illumination Modules}}

To demonstrate CrystalTac's customisation flexibility for prototype designs and its adaptability for modifying mature designs, two bases with different vision and illumination modules are applied to the five proposed sensors. The first customised base is designed for C-Sight and C-SighTac, characterised by a square body with six internal white illumination sources, as detailed in \cite{fan2024design}. The other base is from the well-known Digit \cite{lambeta2020digit}, which features a small black curved housing with an RGB light source. It is a popular commercial VBTS product due to its suitability for mounting on robot hands and its quick-change contact module design. So we used it as the base for C-Tac, Vi-C-Tac, and Vi-C-Sight. A description of these two bases is provided in \textbf{Supplementary \ref{Supplementary 5.3}}.

\section{Performance Evaluation}

In this section, we designed several experiments to evaluate the CrystalTac family across a variety of task targets. Initially, the performance of CrystalTac with different sensing mechanisms was verified through three functional tasks. As C-Sight has been tested on tactile reconstruction task \cite{fan2024design}, it, along with the similar C-SighTac, was not further evaluated. Instead, the focus was on C-Tac, Vi-C-Tac, and Vi-C-Sight, which were selected for object recognition, object and texture hybrid recognition, and see-through-skin exploration, respectively. Additionally, two further evaluations were conducted on the manufacturing cost and customisation flexibility of CrystalTac.

\subsection{Object Recognition}

The C-Tac with single-layer dot markers was used to achieve the object recognition task. This sensor features a black skin to prevent external light interference and a white marker pattern that is sensitive to mapping skin deformation. As illustrated in Fig. \ref{object recognition} (A.a), six print parts were selected, including a dot, ring, sphere, curve, waves and multiple dots. The difference between the reference image and the image after contact can indicate both the location and extent of deformation. For a marker that has been displaced or deformed, its original position is indicated in red, while the final position is indicated in blue; the absence of these colours suggests that the marker remained stationary. For instance, when the hollow ring is in contact, only the peripheral markers of C-Tac are displaced, whereas the centrally located markers are the most displaced when for the sphere.

Subsequently, a total of 1200 images were collected, with 200 images per class, to train the object recognition models. A Densenet121 \cite{huang2017densely} model was employed for data inference, and the evaluation results are summarised in Fig.~\ref{object recognition}(A.b). With a test accuracy of 98.3471\%, the trained model correctly identified 119 out of 121 images in the test set, demonstrating C-Tac's capability to characterise tactile information from physical contact based on the MDM mechanism.

\subsection{Object and Texture Hybrid Recognition}

\begin{figure}[!htbp]
    \centering
    \includegraphics[width = 0.96\hsize]{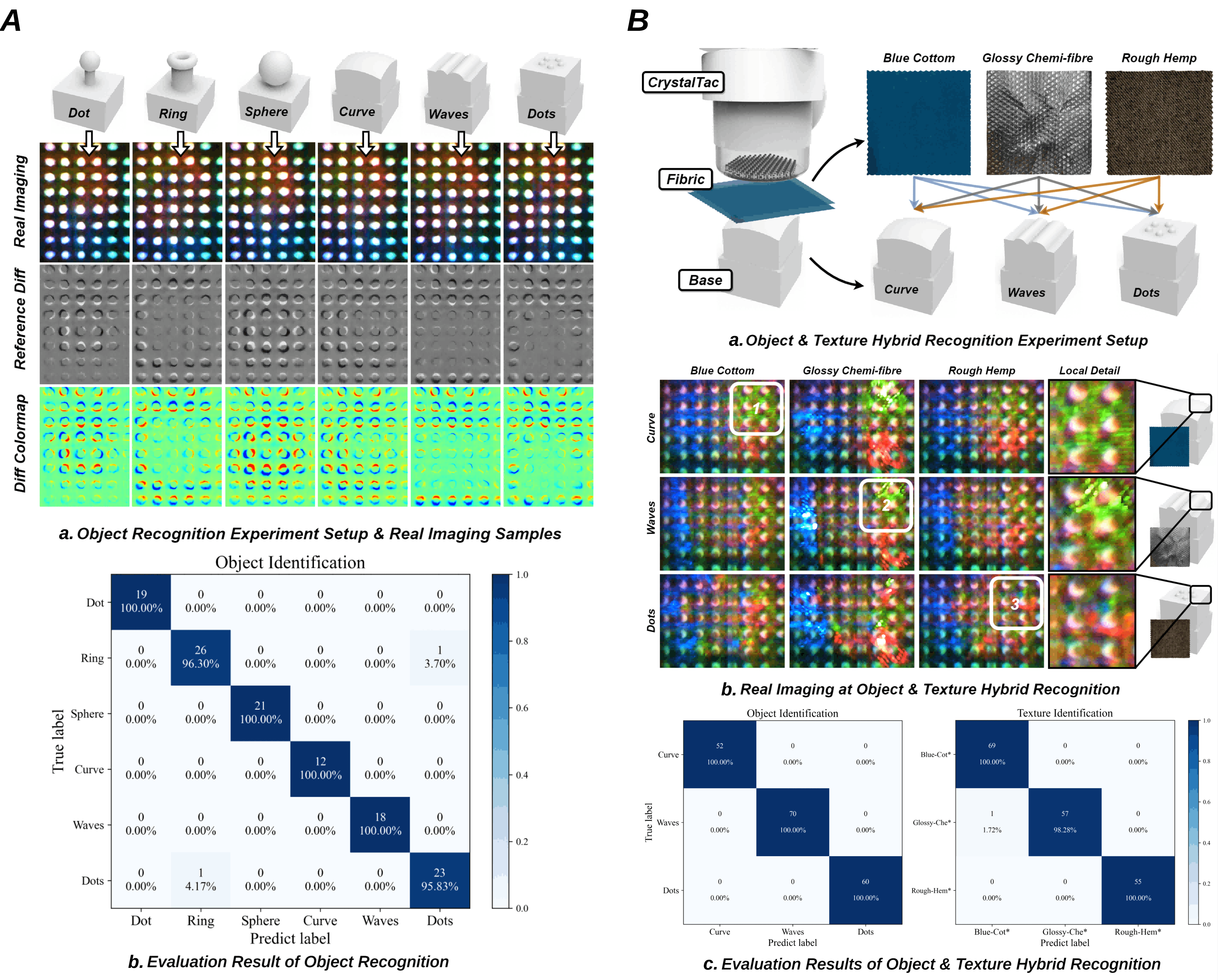}
    \captionsetup{font=small} % Change 'small' to your desired size
    \caption{A: Setup for the object recognition using C-Tac with single-layer dot markers. (a) The marker pattern distributions change with different contact objects. (b) The marker features are effective for precise object identification. B: Setup for the object and texture hybrid recognition using Vi-C-Tac with double-layer markers. (a) By covering the fabric, the sensing of both visual and tactile features can be simultaneously evaluated. (b) Fine textures are clearly visualised, while object shape mapping relies primarily on double-layer marker patterns. (c) Both objects and textures can be identified using Vi-C-Tac.}
    \label{object recognition}
    \vspace{-0.5cm}
\end{figure}

To evaluate the MDM+MFM mechanism provided by Vi-C-Tac, a hybrid recognition experiment was designed, with the setup illustrated in Fig. \ref{object recognition} (B.a). The experiment involved three different fabrics-blue cotton, glossy chemical fibre, and rough hemp-as well as three print parts: curve, waves, and dots. By wrapping the fabric around the objects, visual information alone was insufficient to distinguish the print objects, though it allowed for capturing the fine texture of the fabric. Thereby, the markers were utilised to capture tactile features, enabling the analysis of the shape of the underlying parts. A Vi-C-Tac system with double-layer dot markers was employed. A total of nine permutations between these fabrics and parts were tested, as shown in Fig. \ref{object recognition}(B.b). The enlarged local details revealed distinct textures: cotton fabric displayed clear horizontal grain, chemical fibre exhibited point textures with strong reflective properties, while the features of hemp were less distinct due to its rough characteristics. The double-layer markers were discernible within the view, with non-overlapping parts of the markers differentiated by colour contrast.

For each permutation case, 200 images were collected, resulting in a total of 1800 images for model training. By adding two separate head structures following the output of Densenet121, two classifiers for object recognition and texture recognition were implemented in a decoupled manner. The test results, shown in Fig. \ref{object recognition}(B.c), indicate identification accuracies of 100\% (182/182) for object recognition and 99.45\% (181/182) for texture recognition. Both visual and tactile features were successfully captured by Vi-C-Tac, thanks to the MDM+MFM mechanism.

\subsection{See-through-Skin Exploration}

A Vi-C-Sight was employed to conduct see-through-skin exploration, utilising its IMM+MFM mechanism. The elastomer, made up entirely of Agilus30 Clear, functions as a light-conductive plate \cite{shimonomura2016robotic}, enabling the total internal reflection effect to achieve the fusion of proximity-tactile sensing. Deformed areas are differentiated in brightness from undeformed areas, while other regions remain highly transparent, thereby facilitating the see-through-skin capability. As shown in Fig. \ref{see through skin}(A.a), fine textures on the objects' surfaces are clearly captured. For instance, fingerprints are visible without interference from other marker patterns, allowing Vi-C-Sight to produce such clear imaging that the entire outline of the finger is recognisable against a darker background. A distinct line of demarcation marks the contact areas where textures are mapped onto the elastomer surface, akin to glass. The fabric fibres can embed themselves into the elastomer during contact, enhancing the TIR effect. This effect can be observed in the geometry of criss-cross cotton, the fibre orientation of twill denim, and the level of embroidery deformation, which can all be mapped to variations in pixel intensity. Unlike Vi-C-Tac, the MFM mechanism of Vi-C-Sight inherently couples visual and tactile information through changes in the luminance of the pixels, similar to TIRgel \cite{zhang2023tirgel}.

In Fig. \ref{see through skin}(A.b), an embossed print base with a human figure was wrapped in a skeleton fabric. Vi-C-Sight was used to explore the surface of these combined parts and generate an exploration map. Due to the sparse nature of the fibres, the fabric was not as visually distinctive as the opaque printed adhesive pattern also present. However, the figure pattern of the embossed print base beneath, where the two overlap, is visible in the reddish colour of the embossed areas, indicating a higher degree of deformation and a more pronounced fabric texture in the immediate vicinity.

\subsection{Manufacturing Cost Evaluation of CrystalTac Family}

A similar evaluation of manufacturing costs to \cite{fan2024design} has been applied to the CrystalTac family, with the results summarised in Table \ref{crystaltac sensor cost}. For C-Tac with single-layer dot markers, Vi-C-Tac with double-layer dot markers, and Vi-C-Sight, all of which utilise a Digit base, C-Tac is slightly larger in size and volume due to the additional layer of opaque skin. C-Sight and C-SighTac, which use a customised base, have similar volumes to the others due to their rectangular shape. This difference in shape is reflected in the unit print speed (T/V) and unit material consumption cost (C/V), as C-Tac/Vi-C-Tac/Vi-C-Sight possess complex curved surfaces with overhangs, necessitating additional support materials. This requirement reduces the unit print speed and increases the unit material consumption cost. For instance, the T/V and C/V for C-Tac/Vi-C-Tac/Vi-C-Sight are approximately 11.5 $min/cm^3$ and 0.73 £$/cm^3$, respectively, which is higher compared to C-Sight/C-SighTac, with around 10 $min/cm^3$ and 0.685 £$/cm^3$. Therefore, it can be concluded that the closer the shape of a CrystalTac component is to a rectangle, the more efficient it is to print through monolithic manufacturing, thereby lowering the manufacturing cost.

\begin{table}[!htbp]
\caption{Manufacturing Cost of Different Sensors Within CrystalTac Family.}
\resizebox{\textwidth}{15mm}{
\begin{tabular}{ccccccccccc}
\hline
\textbf{Sensor} & \textbf{Size X/Y/Z($mm$)} & \multicolumn{1}{l}{\textbf{Volume($cm^3$)}} & \multicolumn{1}{l}{\textbf{AG($g$)}} & \multicolumn{1}{l}{\textbf{VR($g$)}} & \multicolumn{1}{l}{\textbf{DG($g$)}} & \multicolumn{1}{l}{\textbf{Sup($g$)}} & \textbf{Time($min$)} & \multicolumn{1}{l}{\textbf{Cost(£)}} & \textbf{T/V($min/cm^3$)} & \textbf{C/V(£/$cm^3$)} \\ \hline
C-Tac           & 34x27x16.5              & 6.538                                    & 9                                   & 12                                & 3                                 & 10                               & 74                 & 4.678                                    & 11.318       & 0.716        \\
C-Sight         & 26.5x26.5x13.5          & 6.446                                    & 10                                  & 10                                & 3                                 & 8                                & 64                 & 4.418                                    & 9.929        & 0.685        \\
C-SighTac       & 26.5x26.5x13.5          & 6.446                                    & 10                                  & 10                                & 3                                 & 8                                & 64                 & 4.418                                    & 9.929        & 0.685        \\
Vi-C-Tac        & 34x27x16.15             & 6.208                                    & 8                                   & 12                                & 3                                 & 10                               & 73                 & 4.478                                    & 11.759       & 0.721        \\
Vi-C-Sight      & 34x27x16.15             & 6.208                                    & 9                                   & 12                                & 3                                 & 9                                & 73                 & 4.608                                    & 11.759       & 0.742        \\ \hline
\end{tabular}}
\label{crystaltac sensor cost}
\noindent{\footnotesize{\textsuperscript{1}  AG, VR, DG and Sup indicate Agilus30 series, Vero series,
DraftGrey, and support. \textsuperscript{2}  cost excludes 20\%VAT.}}
\end{table}

\begin{figure}[!htbp]
    \centering
    \includegraphics[width = 0.96\hsize]{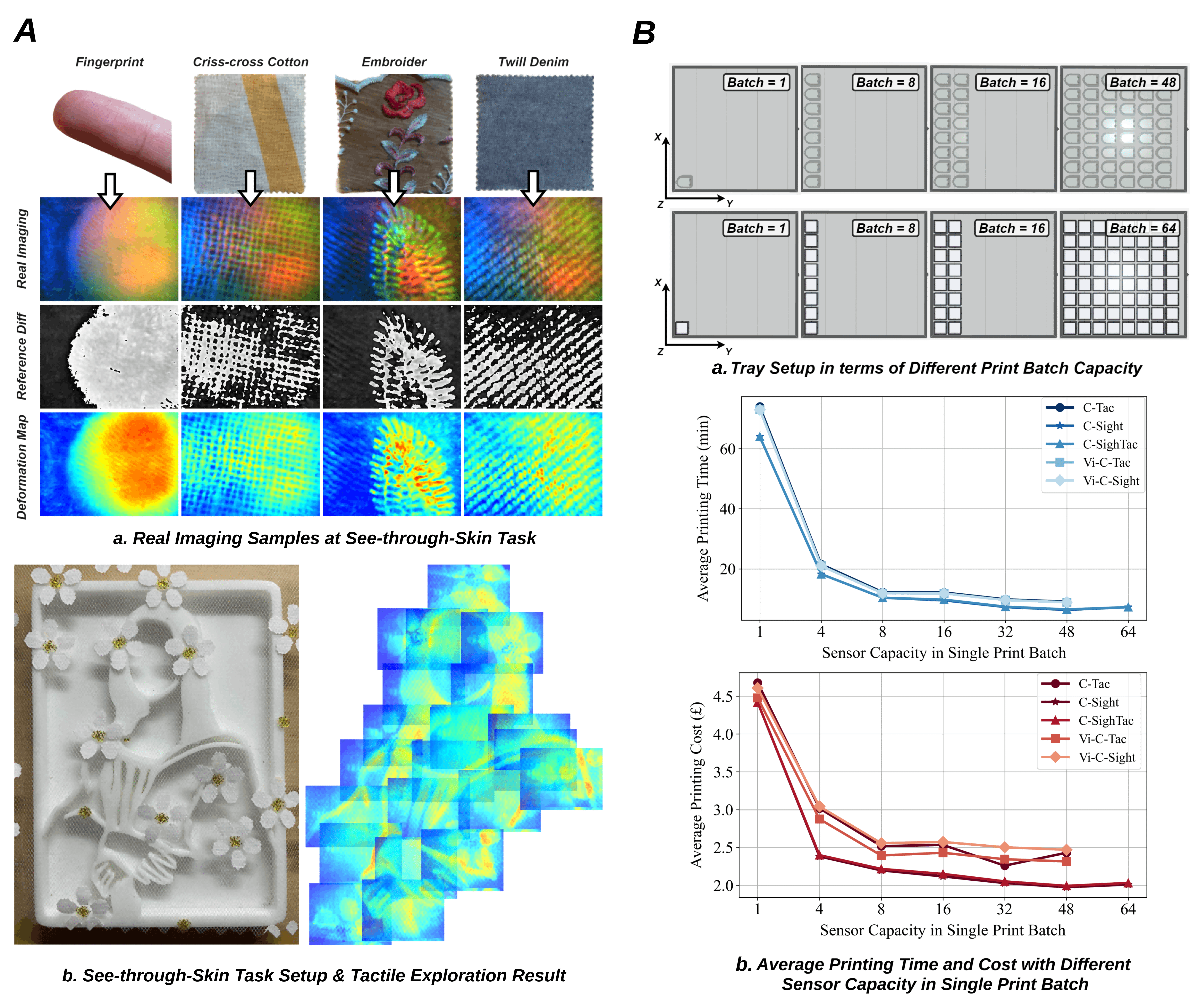}
    \captionsetup{font=small} % Change 'small' to your desired size
    \caption{A: Setup for the see-through-skin exploration using Vi-C-Sight. (a) Pure Agilus30 Clear provides a clear view of the visual-tactile fusion feature. (b) Both the fine texture of the fabric and the geometric shape of the rigid base can be captured. B: Evaluation of CrystalTac's manufacturing cost. (a) The maximum capacity for Crystaltac using the Digit base is 48 units, compared to 64 units using a customised base. (b) As print batch increases, both the manufacturing time and cost decrease gradually until a stable level.}
    \label{see through skin}
\end{figure}

As shown in Fig. \ref{see through skin}(B), further evaluations have been conducted concerning different sensor capacities within a single print batch. For all five types of CrystalTac sensors, the maximum number along the X direction on the print tray is 8, while it varies in the Y direction-6 for C-Tac/Vi-C-Tac/Vi-C-Sight and 8 for C-Sight/C-SighTac-resulting in maximum capacities of 48 and 64, respectively. As the print batch capacity approaches these maximum values, both the average printing time and cost decrease sharply until reaching a batch capacity of 8, after which they stabilise at a slightly lower level. This pattern aligns with the conclusions obtained in \cite{fan2024magictac}. The underlying reason is that when the capacity exceeds 8, additional motion overheads are required in the Y direction for the extra columns. For example, in the case of C-Tac, when the batch capacity is 48, the average print time and cost are only 9.08 minutes and £2.43, representing decreases of 87.73\% and 48.05\% from 74 minutes and £4.678, respectively, when the batch capacity is 1.

\begin{figure*}[!htbp]
    \centering
    \includegraphics[width = 1.0\hsize]{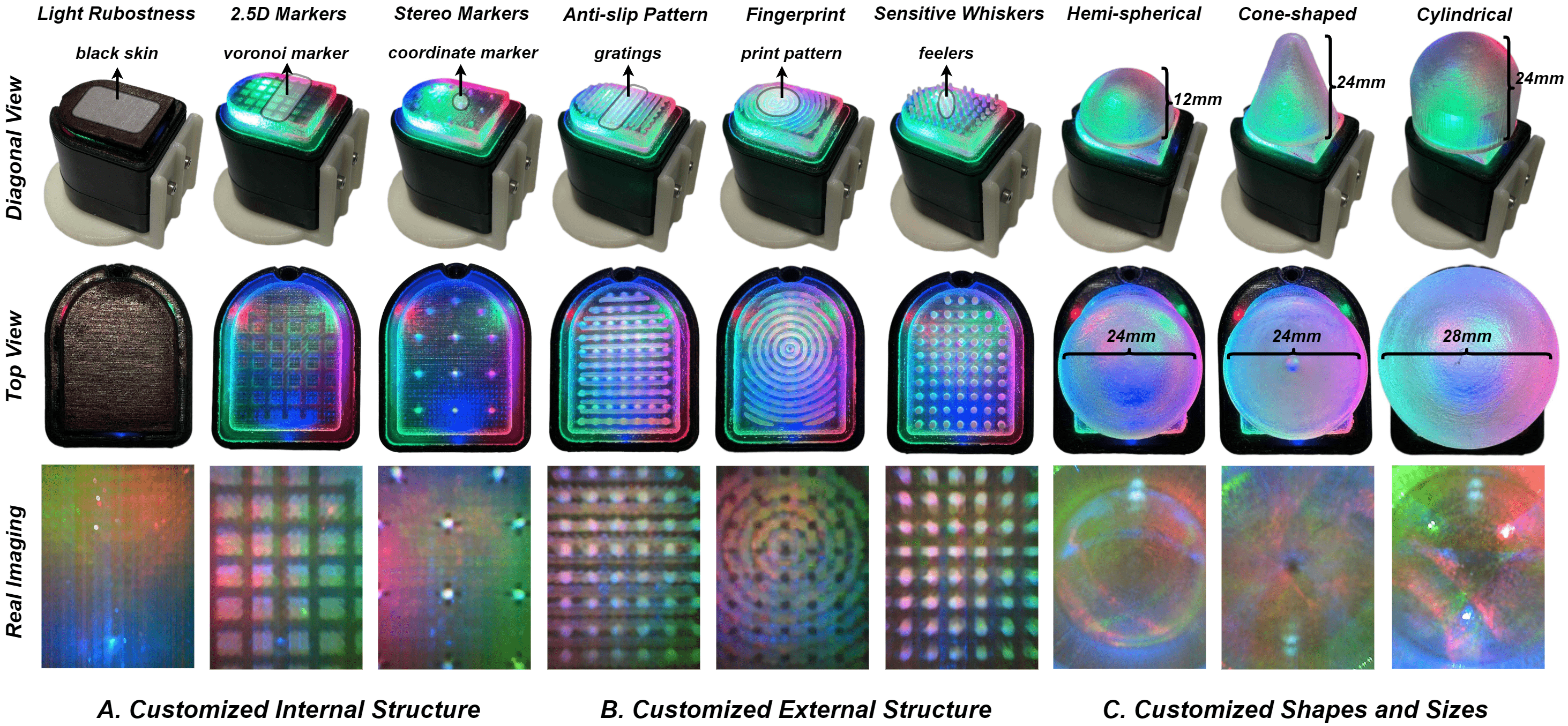}
    \captionsetup{font=small} % Change 'small' to your desired size
    \caption{Customised flexibility test of CrystalTac. A: The internal structure of CrystalTac can be flexibly customised. B: The external accessories are free to add. C: The scales and shapes are also customisable.}
    \label{customised flexibility}
\end{figure*}

\subsection{Customised Flexibility Evaluation of CrystalTac Family}

As discussed, customised flexibility is an essential metric for reducing design complexity. With monolithic manufacturing, as long as the CAD model of CrystalTac is designed, it can be manufactured directly without any additional processes. This capability allows for the realisation of many complex designs to meet the requirements of specific tasks.

\subsubsection{Customised Internal Structure} As shown in Fig. \ref{customised flexibility}(A), for designs that only require tactile information, light robustness is enhanced by using Agilus30 Black for the outer skin, which blocks external ambient light. Additionally, the 2.5D markers made of soft material can deform to provide richer tactile information. In contrast, markers made of rigid material maintain their structure while still moving with deformation, enabling the creation of more complex geometries. Further design examples of novel markers and complicated structures can be seen in \textbf{Supplementary \ref{Supplementary 5.3}} and \textbf{Supplementary \ref{Supplementary 5.4}}.

\subsubsection{Customised External Structure} The external structure plays a crucial role in tactile sensing, particularly in dynamic interactions. As shown in Fig. \ref{customised flexibility}(B), the anti-slip texture can increase friction between the sensor surface and external objects, effectively reducing undesirable sliding and providing a solid foundation for robust gripping tasks. This texture also protects the contact module body by being wear-resistant, thus extending the service life and reducing usage costs. Furthermore, whiskers can enhance the sensitivity of built-in markers. By leveraging the principle of micro-leverage effect, tactile information such as shear or pressure received at the external end is magnified at the internal end.

\subsubsection{Customised Shapes and Sizes} The ability to modify the size and shape of the contact module is also a key feature. Compared to the traditional flat shape, spherical and cylindrical surfaces perform better in in-hand manipulation tasks since they can obtain more sensing area. However, these surfaces are more difficult to manufacture, especially internal lenses, which often require complicated processes that challenge traditional manufacturing methods. As shown in Fig. \ref{customised flexibility}(C), a hemispherical sample and two additional samples with cone-shaped and cylindrical structures were manufactured. Their built-in lenses are consistent with the overall shapes, with elastomers all 2mm thick, matching the lens thickness.

\section{Conclusions and Future Work}

In this work, we introduce a sensor family named CrystalTac, comprising five branches with unique designs. Building on the previous work of C-Sight, the monolithic manufacturing technique has demonstrated its capability to address the existing challenges in the traditional design and creation process, showing potential to become a universal manufacturing technology for VBTSs. Accordingly, we summarised the sensing mechanisms of currently known VBTSs, including IMM, MDM, MFM, and multi-mechanism fusion. Based on these typical mechanisms, five types of CrystalTac sensors were fabricated using monolithic manufacturing: C-Tac, C-Sight, C-SighTac, Vi-C-Tac, and Vi-C-Sight. Subsequent functional experiments demonstrated that the CrystalTac sensors perform well and meet their design targets with unique tactile sensing mechanisms. Additionally, monolithic manufacturing has led to significant improvements in manufacturing costs and design flexibility, often constrained by conventional methods of manufacturing and assembly.

The CrystalTac family can be regarded as an initial template, with no strict parameter limitations for each sensor detail, thereby encouraging the community to view it as a foundation for further development. The significance of this work lies in demonstrating the capability of monolithic manufacturing to produce VBTSs with various tactile sensing mechanisms, providing confidence and inspiration to other researchers in the tactile robotics field. In future work, we aim to advance the new CrystalTac series by enhancing the capabilities of monolithic manufacturing in terms of production quality and efficiency, as well as multi-material printing for VBTSs. These technologies can be seamlessly integrated with tactile sensory enhancements in dexterous hands to perform tasks such as human-computer interaction or dexterous manipulation.

\section*{Acknowledgment}
The authors would like to thank Andrew Stinchcombe, Ugnius Bajarunas, and Tom Barnes for their assistance and valuable advice in technique development and hardware fabrication.

\subsection*{Author Contributions} 
Conceptualisation, W.F. and H.L.; hardware, W.F. and H.L.; software, W.F.; writing---original draft preparation, W.F.; writing---review and editing, H.L. and D.Z.; visualisation, W.F. and H.L.; supervision, D.Z.; project administration, D.Z. All authors have read and agreed to the published version of the manuscript.

\subsection*{Funding}
Wen Fan and Haoran Li are partially funded by the CSC scholarship. The authors would like to acknowledge the Royal Society Research Grant (RGS/R1/221122).

\subsection*{Conflicts of Interest}

The author(s) declare(s) that there is no conflict of interest regarding the publication of this article.

\subsection*{Data Availability}

Please contact the authors to obtain data, including the design details of the CrystalTac family.

\printbibliography

\newpage

\section*{Supplementary Materials}

\subsection{Optimisation on Sub-components Manufacturing of CrystalTac}
\label{Supplementary 5.1}

Several extended experiments have been conducted to explore the optimal range of sub-component attributes in CrystalTac, including the lens and elastomer.

\begin{figure}[!htbp]
    \centering
    \includegraphics[width = 0.7\hsize]{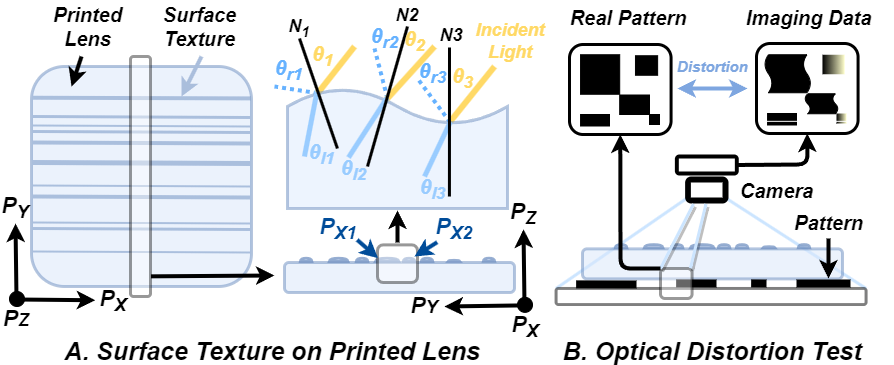}
    \captionsetup{font=small} % Change 'small' to your desired size
    \caption{A: Horizontal textures on the printed lens surface. Adjacent textures form a wavy pattern, complicating the refraction and reflection of incident light. B: These surface textures can result in unpredictable distortion between the actual pattern and the camera imaging.}
    \label{lens distortion}
\end{figure}

\subsubsection{\textbf{Optimisation for CrystalTac Lens Manufacturing}}

VeroClear is highly effective for manufacturing lens components, being a nearly colourless material that exhibits dimensional stability. Its properties are comparable to those of Polymethyl methacrylate (PMMA), commonly known as acrylic. Its upgraded version, VeroUltraClear, offers 95\% light transmission and improves upon VeroClear with higher clarity, transparency, and a lower yellow index. We produced a batch of lens samples using VeroClear, which demonstrated good transparency. However, there were still strip-shaped textures present on the surface of the printed lenses, as depicted in Fig. \ref{lens distortion}(A).

The potential cause of these textures could be the alternate superimposition of the axial movements Px and Py of the print head in the X/Y direction. When two adjacent movements, Px1 and Px2, separated by $\Delta$Py, produce a deviation $\Delta$Pz in the vertical height Pz possibly due to a mechanical error or the state of material polymerisation-an uneven texture is created between the Px1 and Px2 trajectories. These textures complicate the processes of refraction and reflection as incident light passes through the material. Assuming that the normal incidence angles of three parallel beams on the lens surface are N1, N2, and N3, with angles of incidence $\theta$1, $\theta$2, and $\theta$3, respectively, the angles of reflection for their reflected light will be $\theta$r1, $\theta$r2, and $\theta$r3. It is evident that these reflected beams are no longer parallel. Similarly, for light refraction, which follows Snell's law as shown in Eq. \ref{snell law}, the refractive indices ${N_{lens}}/{N_{air}}$ of the two media (lens and air) are greater than 1. Therefore, the refraction angles $\theta$l1, $\theta$l2, and $\theta$l3 will be slightly smaller than $\theta$1, $\theta$2, and $\theta$3, and these refracted beams will also not be parallel to each other.

\begin{equation}
\label{snell law}
\frac{\sin \theta}{\sin \theta l} = \frac{N_{lens}}{N_{air}} 
\end{equation}

When applied to CrystalTac, the aforementioned phenomenon of printed lenses leads to two problems. Firstly, internal illumination can cause irregular reflections on the lens surface. Secondly, images captured through the lens may exhibit optical distortion, as demonstrated in Fig. \ref{lens distortion}(B). Both issues significantly impair the quality of the tactile data acquired by CrystalTac. The first problem can be addressed by optimising the design of the illumination system. The second issue, however, differs from common optical lens distortions such as barrel distortion, pincushion distortion, and mustache distortion, which can be corrected using distortion parameters derived from chessboard pattern calibration. This optical distortion arises from the randomly shaped surface of the lens, leading to unpredictable outcomes that cannot be corrected through simple parameter adjustments.

\begin{figure}[!htbp]
    \centering
    \includegraphics[width = 1\hsize]{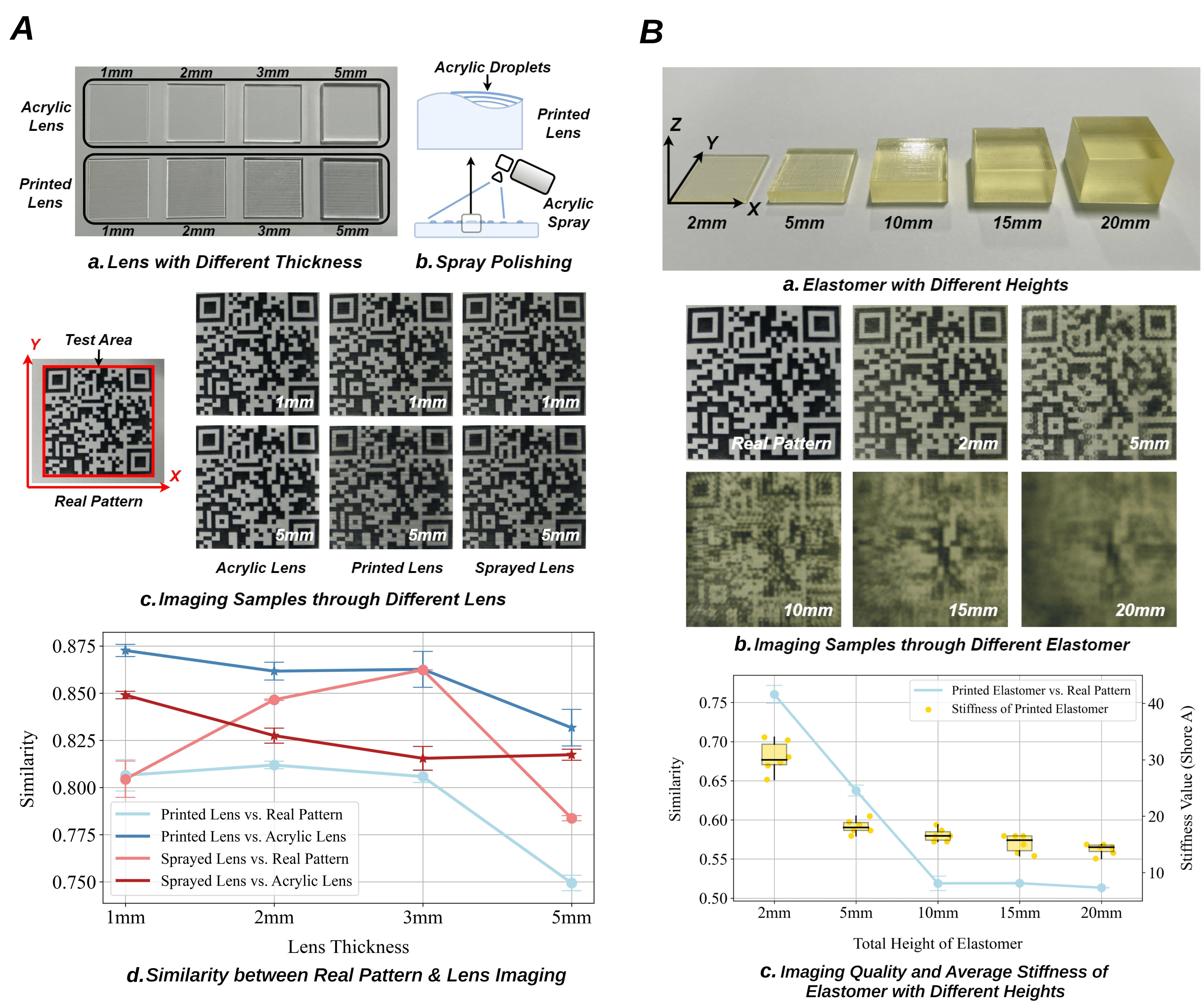}
    \captionsetup{font=small} % Change 'small' to your desired size
    \caption{A: Optimisation of printed lens for CrystalTac. (a) Test samples of printed lenses and acrylic lenses were compared; (b) The acrylic spray can enhance the glossy of the printed lens; (c) Real imaging for evaluating lens quality; (d) Imaging quality test of the lenses was conducted, with the error value indicating the difference between similarity values in X/Y directions. B: Optimisation of printed elastomer for CrystalTac. (a) Test samples of printed elastomer with different heights were compared; (b) Real imaging for evaluating elastomer quality; (c) Trend of imaging quality and stiffness with elastomer thickness change.}
    \label{lens evaluation}
\end{figure}

This issue is evidently a consequence of the operating principles of pp, which, despite having a surface finish superior to most other 3D printing technology. To avoid redundant post-processes such as mechanical polishing, two methods have been tested to determine if the imaging quality of the printed lens surface can be improved:

\begin{itemize}

\item \textbf{Adjusting Printing Thickness}:

As shown in Fig. \ref{lens evaluation}(A.a), four lenses with different thicknesses were printed: 1mm, 2mm, 3mm, and 5mm. All lenses have planar dimensions of 30mm x 30mm. For the control group, four additional acrylic lenses of the same size were fabricated using laser cutting. As indicated in Fig. \ref{lens evaluation}(A.c), by placing a QR code pattern beneath the lenses, a camera positioned above can capture the pattern's imaging through the lenses. To quantitatively analyse the optical distortion level, the Structural Similarity Index Measure (SSIM) was introduced as the metric to evaluate image similarity. This approach assesses similarity by considering a combination of factors, including image luminance, contrast, and structure. To avoid the influence of surface texture direction, each printed lens was placed over the QR pattern along both the X and Y directions, capturing two image samples. From Fig. \ref{lens evaluation}(A.d), the similarity between images taken through printed lenses and acrylic lenses is around 86\% for thicknesses less than 3mm. However, as thickness increases to 5mm, similarity decreases to 83\%. This decrease is primarily attributed to the lower light transmittance of VeroClear compared to acrylic, with excessive thickness resulting in darker images through the printed lenses. The highest similarity score, approximately 87.5\%, is attained with lenses 1mm thick, suggesting that some distortion may be due to surface texture. Similarly, when comparing the similarity to the actual pattern, there is an overall decrease of about 7\%, indicating that acrylic lenses also contribute to some degree of distortion.

\item \textbf{Applying Acrylic Spray}: 

As illustrated in Fig. \ref{lens evaluation}(B), applying acrylic spray can fill the uneven texture on the lens surface with liquid droplets. Once these droplets are cured, which can enhance the overall finish of the lens, offering a more convenient and efficient alternative to physical polishing using sanding equipment. In our test, all four printed lenses were placed together and a canister acrylic spray was applied 1~m above them for around 2~s. After waiting for an hour under ventilated conditions, the liquid acrylic layer on their surface was fully cured and ready for testing, with all the settings being the same as in the previous. Also from Fig. \ref{lens evaluation}(D), the similarity between sprayed lens imaging and acrylic lens imaging is down overall by 3-5\% compared to printed lens imaging without post-process. However, the similarity to the real pattern has been improved, reaching a maximum of 86\% with lenses 3mm thick. The improvement in the impact of surface texture due to the application of the spray is significant, as indicated by the reduced disparity in similarity values across the X and Y directions for lenses of all thicknesses.

As illustrated in Fig. \ref{lens evaluation}(A.b), applying acrylic spray can fill the uneven texture on the lens surface with liquid droplets. Once cured, these droplets enhance the overall finish of the lens, offering a more convenient and efficient alternative to physical polishing with sanding equipment. In our test, all four printed lenses were placed together, and a canister of acrylic spray was applied from a distance of 1m for approximately 2 seconds. After curing for an hour in ventilated conditions, the liquid acrylic layer on the surface was ready for testing, with all settings identical to the previous tests. As shown in Fig. \ref{lens evaluation}(A.d), the similarity between images taken through sprayed lenses and acrylic lenses decreased overall by 3-5\% compared to images from printed lenses without post-processing. However, the similarity to the real pattern improved, reaching a maximum of 86\% with 3mm thick lenses. The improvement in the impact of surface texture due to the spray application is significant, as indicated by the reduced disparity in similarity values across the X and Y directions for lenses of all thicknesses.

\end{itemize}

In summary, altering the printing thickness and applying acrylic spray both impact the imaging quality of printed lenses. Considering various factors, a thickness range of 2-3mm is optimal for printed lenses, providing a balance between structural strength and satisfactory imaging performance. Building on this foundation, the application of acrylic spray improves the surface finish of the printed lens.

\subsubsection{\textbf{Optimisation for CrystalTac Elastomer Manufacturing}}

Most VBTSs require a transparent, silicone-like elastomer as the core material to facilitate technologies such as IMM, MDM, MFM, and multi-mechanism fusion. Agilus30 Clear, due to its transparent and flexible texture, is suitable for replacing silicone. However, the Shore hardness of pure Agilus30 is 30A, which is too hard for designs requiring a softer elastomer. For instance, GelSight typically uses a Shore hardness range of 5-20A \cite{yuan2017gelsight}. By incorporating the multi-layer grid structure proposed in MagicTip \cite{fan2024magictac}, the stiffness of the printed elastomer can be further reduced. To assess the impact of this structure on imaging quality, five test samples were manufactured. These samples shared the same X/Y dimensions of 30mm x 30mm for the elastomer, but varied in height: 2mm, 5mm, 10mm, 15mm, and 20mm, as shown in Fig. \ref{lens evaluation}(B.a). The skin layer's thickness was set to 0.5mm to balance a soft texture with stable strength. The same QR code pattern and SSIM metric used for evaluating the imaging quality of printed lenses, as depicted in Fig. \ref{lens evaluation}(A.b), were employed.

As displayed in Fig. \ref{lens evaluation}(B.b), as the height of the printed elastomer increases, the imaging quality decreases in two ways. Firstly, the image tone gradually becomes darker and tends toward a darker yellow. This is primarily due to the support material filled in the internal core, which has lower light transmission than the Agilus30 material. Secondly, the imaging quality deteriorates, leading to a gradual loss of detail in the pattern beneath. This trend is confirmed by Fig. \ref{lens evaluation}(B.c), where the similarity between the printed elastomer imaging and the real pattern is 76\% at a height of 2mm but gradually drops to around 52\% when the height exceeds 10mm. We also found that overall height affects stiffness to a certain extent, even with the same outer skin thickness, particularly when the height is less than 5mm. This is mainly because a reduction in elastomer height reduces the number of layers in the embedded grids, and too few layers limit the range of deformation that can occur in the support core.

For CrystalTac, to optimise the imaging quality and elasticity of the printed elastomer with the multi-layer grid structure, the suitable height ranges are 2mm to 5mm, where thinner layers lead to better transparency and thicker layers result in softer stiffness.

\subsection{Marker Design for CrystalTac Family}
\label{Supplementary 5.2}

\begin{figure}[!htbp]
    \centering
    \includegraphics[width = 1\hsize]{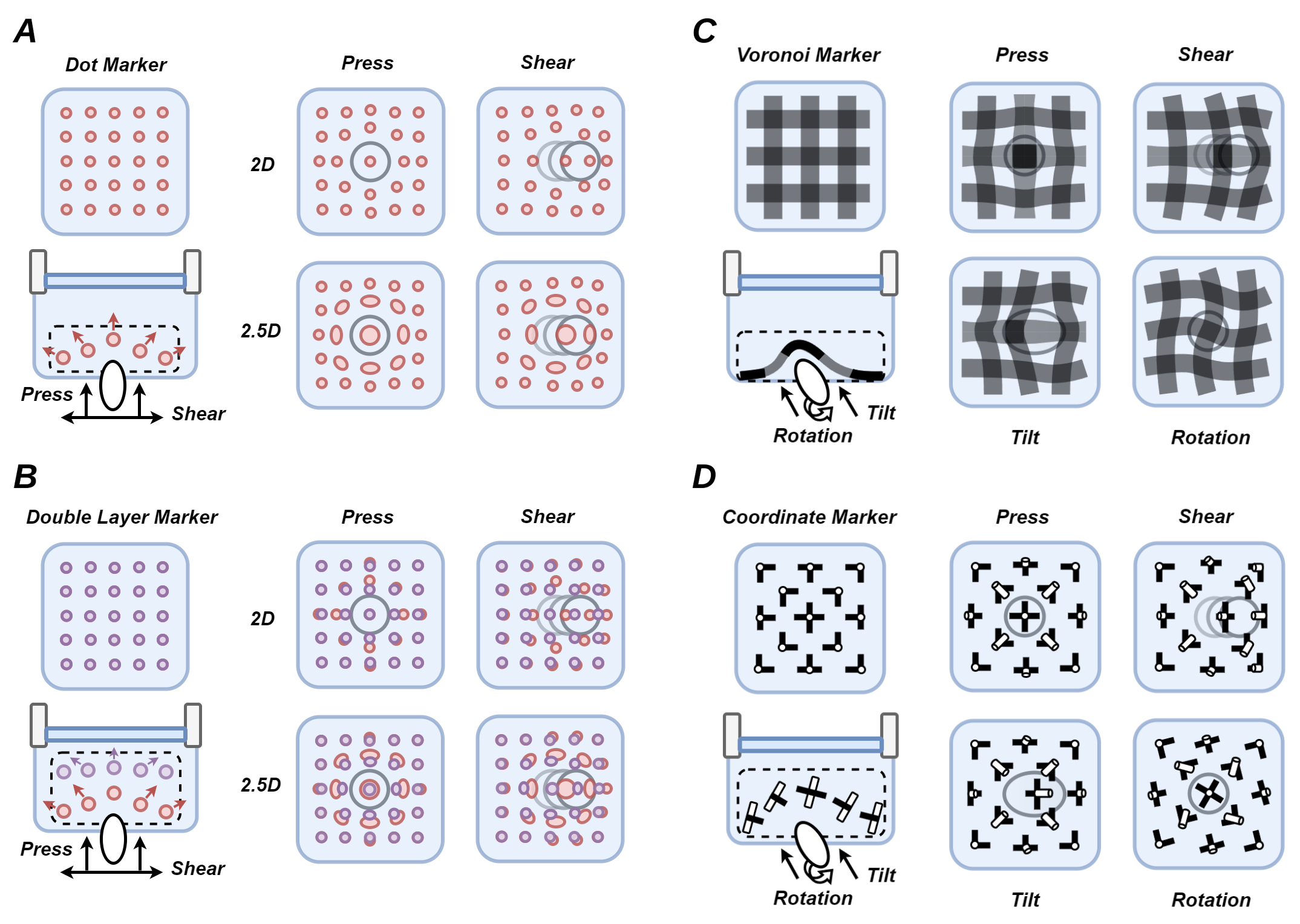}
    \captionsetup{font=small} % Change 'small' to your desired size
    \caption{Marker design for CrystalTac. A: Schematic of dot marker. The pattern of dot markers is sensitive to translations along the contact plane. Compared to 2D dot markers, 2.5D dot markers can change their shapes to provide more features about depth. B: Double-layer markers enrich tactile features through changes between the upper and lower-layer patterns. 2.5D double-layer markers extend this capability from geometric distribution to shape variety. C: The Voronoi marker represents a geometric advancement over the 2.5D dot marker. These regularly arranged geometric tessellations use variations in the continuity of corner points, edges, colour, and area of each inlay to map tactile features. D: Each coordinate marker is designed as a gyroscope-like 3D pointer structure, allowing each marker to individually map 6D pose information. This design improves the tactile representation, aiding in force estimation or dynamic motion tracking.}
    \label{dot marker}
\end{figure}

\begin{itemize}

\item \textbf{Dot Marker}: 
Dot markers are among the most common patterns in VBTSs \cite{sato2009finger, taylor2022gelslim, kim2022uvtac}. Regarding shape, dot markers can include both cylinders with slender thicknesses and spheres with symmetrical dimensions. The overall marker array distribution can be categorised into either uniform or random spatial arrangements. Furthermore, dot markers can be transformed into pin-shaped markers by integrating them with the skin, a feasibility demonstrated in \cite{lepora2021soft, ward2018tactip, lepora2022digitac}. As shown in Fig. \ref{dot marker}(A), if a rigid material such as Vero is used for dot marker manufacturing, the printed pattern will exhibit properties of a 2D marker. However, if flexible materials like Agilus30 or Tango are used, the markers have the potential to deform into various shapes. This deformation is an additional feature, categorising them within the realm of 2.5D markers. The precise pattern of the markers is dictated by the design specifications established during CAD modelling, which involves considerations of both the number and density distribution of markers, allowing for the customisation of either a uniform or a random distribution.

\item \textbf{Double-layer Marker}: When the elastomer contacts the object, the extent of elastic deformation decreases along the longitudinal axis. The double marker approach utilises this inhomogeneity by capturing the difference in distribution between the upper and lower pattern layers to map tactile information \cite{scharff2022rapid}. As shown in Fig. \ref{dot marker}(B), this contrasting disparity typically enhances specific dynamic tactile characteristics, such as shear motion, compared to a single-layer marker configuration. The fabrication process for double-layer markers is similar to that of dot markers, with the primary distinction being the inclusion of an additional marker layer in the CAD modelling. Moreover, the construction principles of spatial structures comprising more than two layers remain consistent. By increasing the number of layers, the extreme case is that the different layers of the marker transition from a discrete state of separation to a continuous state of interconnection, resembling a top-to-bottom cylinder. By controlling the gradient distribution of colours, it should be possible to provide a more fine-grained mapping of the tactile information. All of the above variants can be easily implemented through monolithic manufacturing.

\item \textbf{Voronoi Marker}:  Similar to the lattice pattern in \cite{nagata1999feature}, the Voronoi marker is designed with a continuous pattern distribution in a square layout as shown in Fig. \ref{dot marker}(C). Each edge of such a square marker can be viewed as a collection of dot markers, thus improving spatial resolution compared to the standard dot marker, as investigated in continuous marker pattern (CMP) \cite{li2023improving}. Using monolithic manufacturing, Voronoi markers can be seamlessly integrated into the fabrication of the skin or embedded at any specific spatial layer within the elastomer, according to a customised design. However, the marker design must adhere to geometric principles, where each tessellation cell of the Voronoi pattern should be a polygon, such as a triangle, square, or hexagon.

\item \textbf{Coordinate Marker}: Most marker patterns share two characteristics: each marker typically has a regular spatial shape, such as a sphere or dot, and multiple markers are grouped into an array to map force, shear, or other contact information. These characteristics imply that a single marker plays a limited role in the overall array due to the relatively low marker density. Through monolithic manufacturing, we introduce the coordinate marker, as illustrated in Fig. \ref{dot marker}(d). Each individual marker in the array can independently infer the 6D pose of its localised elastomer. This capability stems from the distinct design of each marker, akin to a three-dimensional pointer within a gyroscope, making it sensitive to both rotational and translational movements. Consequently, markers characterised by complex spatial structures, as opposed to regular shapes, can be classified as stereo markers(including proposed coordinate markers). To our knowledge, the design and creation of such stereo markers are achievable exclusively through monolithic manufacturing.

\end{itemize}

\subsection{Vision and Illumination Design for CrystalTac Family}
\label{Supplementary 5.3}

\begin{figure}[!htbp]
    \centering
    \includegraphics[width = 1\hsize]{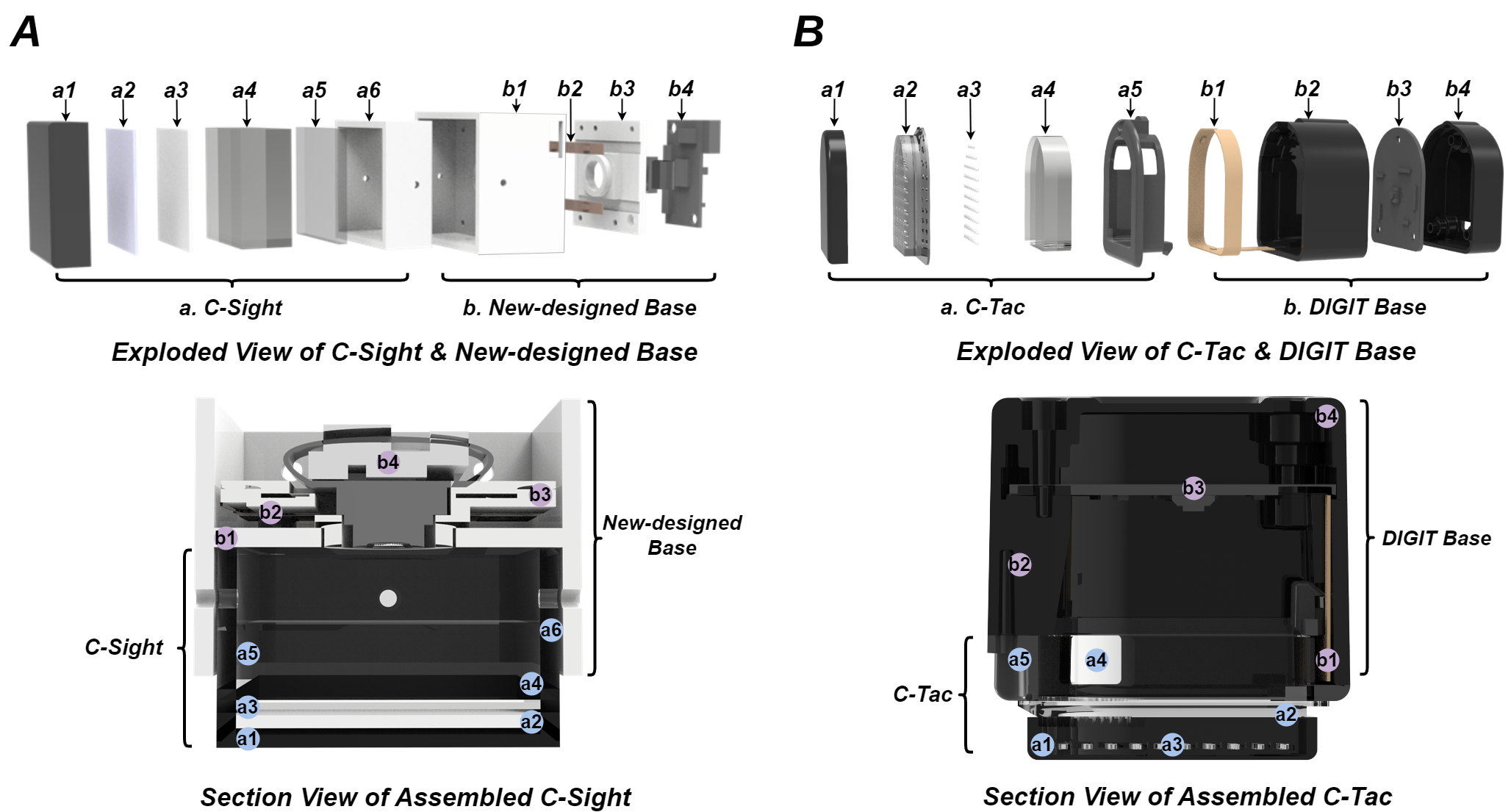}
    \captionsetup{font=small} % Change 'small' to your desired size
    \caption{Base design for proposed CrystalTac, the exploded view, and section view of assembled C-Sight and C-Tac are introduced here for reference. A: New-designed base \cite{fan2024design} for C-Sight and C-SighTac, (a1)black skin; (a2)translucent gel; (a3)white elastomer; (a4)clear elastomer; (a5)lens; (a6)base of contact module; (b1)mounting base; (b2)LED; (b3)camera base; (b4)camera. B: Commercial Digit base for C-Tac, Vi-C-Tac and Vi-C-Sight, (a1)black skin; (a2)clear elastomer; (a3)marker; (a4)lens; (a5)base of contact module; (b1)LED; (b2)mounting base; (b3)camera; (b4)Bottom base.}
    \label{base}
\end{figure}

\subsection{Printing Sample of Customised CraystalTac}
\label{Supplementary 5.4}

\begin{figure}[!htbp]
    \centering
    \includegraphics[width = 1\hsize]{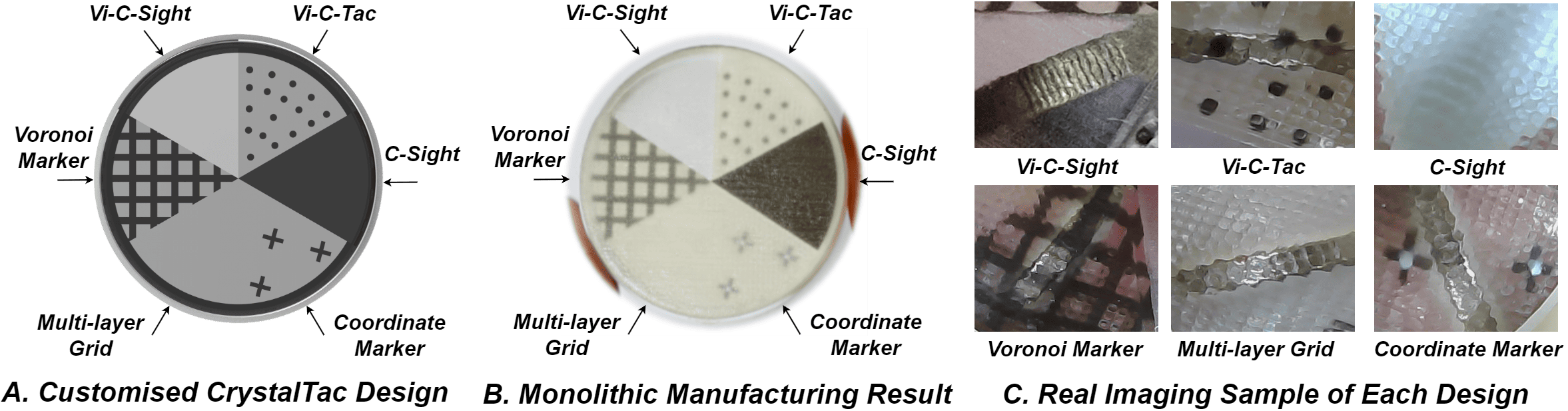}
    \captionsetup{font=small} % Change 'small' to your desired size
    \caption{Monolithic manufacturing provides CrystalTac family superior design flexibility and creation efficiency. A: The CAD model of customised CrytalTac example which integrates six different designs, which is hard to be fabricated by traditional methods. B: The real product of such complicated design through monolithic manufacturing.  C: The images of different customised designs with a coin as stimuli.}
    \label{base}
\end{figure}

\end{document}